\documentclass[lettersize,journal]{IEEEtran}
\usepackage{amsmath,amsfonts}
\usepackage{algorithmic}
\usepackage{algorithm}
\usepackage{array}

\usepackage[caption=false,font=normalsize,labelfont=sf,textfont=sf]{subfig}
\usepackage{caption}
\usepackage{textcomp}
\usepackage{stfloats}
\usepackage{url}
\usepackage{verbatim}
\usepackage{graphicx}
\usepackage{cite}
\usepackage{booktabs} 
\usepackage{threeparttable}
\usepackage{multirow}
\usepackage[colorlinks,linkcolor=blue]{hyperref}
\usepackage{orcidlink}
\usepackage{colortbl}
\usepackage{doi}

\begin{document}

\title{Multi-modal Attribute Prompting for Vision-Language Models}

\author{Xin Liu~\orcidlink{0000-0002-3791-3984}, 
    Jiamin Wu, 
    Wenfei Yang\textsuperscript{$^\dagger$ }
    Xu Zhou,
    Tianzhu Zhang\textsuperscript{$^\dagger$ \thanks{$^\dagger$Corresponding author.}}\orcidlink{0000-0003-0764-6106}
    %
    \thanks{Xin Liu, Jiamin Wu, Wenfei Yang, and Tianzhu Zhang are with the School of Information Science and Technology, University of Science and Technology of China, Hefei 230027, China (e-mail: 
    xinliu99@mail.ustc.edu.cn;
    jiaminwu@mail.ustc.edu.cn; yangwf@ustc.edu.cn; tzzhang@ustc.edu.cn).}    

    \thanks{Xu Zhou is with the Sangfor Technologies Inc., Shenzhen 518000, China (e-mail: zhouxu@sangfor.com.cn).}

\thanks{Copyright © 2024 IEEE. Personal use of this material is permitted. However, permission to use this material for any other purposes must be obtained from the IEEE by sending an email to pubs-permissions@ieee.org.}
\thanks{The definitive version of this paper can be found at:~\href{https://doi.org/10.1109/TCSVT.2024.3424566}{10.1109/TCSVT.2024.3424566}}
}

 
\markboth{IEEE TRANSACTIONS ON CIRCUITS AND SYSTEMS FOR VIDEO TECHNOLOGY,~Vol.~XX, No.~XX, May~2024}%
{Shell \MakeLowercase{\textit{Liu et al.}}: Multi-modal Attribute Prompting for Vision-Language Models}


\maketitle

\begin{abstract}
Pre-trained Vision-Language Models (VLMs), like CLIP, exhibit strong generalization ability to downstream tasks but struggle in few-shot scenarios.
Existing prompting techniques primarily focus on global text and image representations, yet overlooking multi-modal attribute characteristics. This limitation hinders the model's ability to perceive fine-grained visual details and restricts its generalization ability to a broader range of unseen classes. To address this issue, we propose a Multi-modal Attribute Prompting method (MAP) by jointly exploring textual attribute prompting, visual attribute prompting, and attribute-level alignment. The proposed MAP enjoys several merits. First, we introduce learnable visual attribute prompts enhanced by textual attribute semantics to adaptively capture visual attributes for images from unknown categories, boosting fine-grained visual perception capabilities for CLIP. Second, the proposed attribute-level alignment complements the global alignment to enhance the robustness of cross-modal alignment for open-vocabulary objects. To our knowledge, this is the first work to establish cross-modal attribute-level alignment for CLIP-based few-shot adaptation. Extensive experimental results on 11 datasets demonstrate that our method performs favorably against state-of-the-art approaches.
\end{abstract}


\section{Introduction}
\IEEEPARstart{P}{re-trained}  Vision-Language Models (VLMs), such as CLIP~\cite{CLIP} and  ALIGN~\cite{ALIGN}, have demonstrated promising generalization power and transferability on a wide range of downstream tasks~\cite{mei2022guest,zhang2020language,wei2024fine,zhu2023esa,zhou2024unsupervised,lin2024clipose,wang2023tridentcap}, including image
classification~\cite{CLIP}, object detection~\cite{vlm-adapt2,vlm-adapt3}
and 3D understanding~\cite{3dscene,3d-2,3d-3}.
Through contrastive training on a large-scale dataset of image-text pairs, CLIP achieves a global alignment between images and textual descriptions by learning a joint embedding space. The robust cross-modal alignment empowers the CLIP model with the open-vocabulary visual recognition capability. 
In CLIP, class-specific weights for open vocabulary classification can be constructed by plugging the \textbf{class name} in a predefined prompt template like `A photo of a [CLASS].'
Despite its impressive generalization capability, it remains challenging to adapt CLIP to downstream tasks in few-shot scenarios.  
{Due to the large number of parameters in CLIP and the limited number of samples in few-shot task settings, naive fine-tuning of the entire model would likely lead to overfitting, resulting in performance degradation~\cite{clip-adapter, CoOp}.}

 \begin{figure*}[t]
  \centering
  \scalebox{0.8}{  \includegraphics[width=0.99\linewidth]{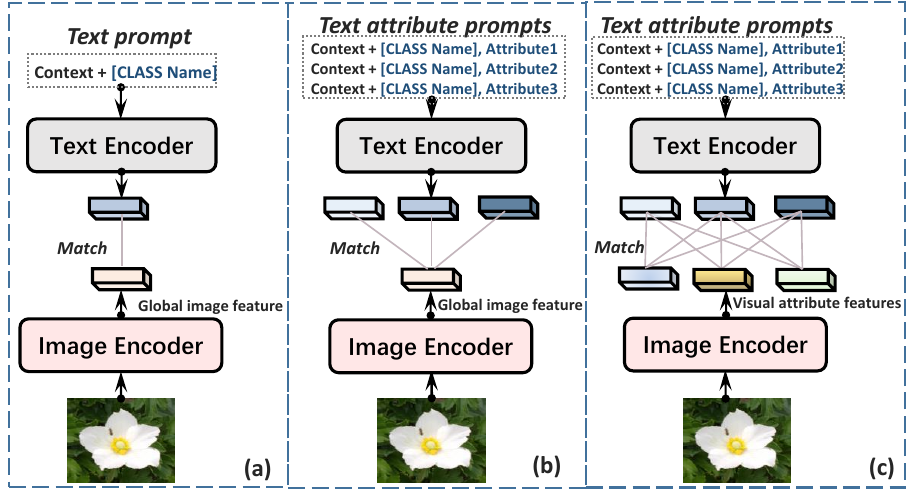}}

   \caption{(a) Conventional prompting methods use hand-crafted or learnable context in combination with the class name to construct the text prompt. 
   (b) Recent methods introduce attribute descriptions to create text attribute prompts containing more semantic content. (c) Our method jointly explores multi-modal attributes and attribute-level alignment, enhancing fine-grained visual perception and achieving attribute-level alignment between images and text categories. }

   \label{fig:contra}
\end{figure*}

%
To enhance the few-shot adaptation capability of CLIP, prompting techniques~\cite{SubPT,cocoop,dapt,PromptSRC,ProDa,rpo,Maple}, such as CoOp~\cite{CoOp} and CoCoOp~\cite{cocoop} have been proposed. 
 These techniques replace hard template context with learnable context in combination with the class name to construct the text prompt. The classification result can be obtained by calculating the similarity between the global image feature and the encoded text prompt. However, as shown in Figure \ref{fig:contra} (a), these prompting methods rely solely on class names and may struggle to fully encapsulate categorical semantics when new unseen classes emerge, causing an issue of `lexical weak tie' where the class name has a tenuous link with its literal semantics. Consider `Rocky Road' as an example, which textually resembles `rock' and `road' but refers to a dessert in reality. When introduced as a new class, the classification weight generated by the model may diverge from its true semantics, potentially causing misclassification. To address this issue, recent works~\cite{Multi-des,classification_via_des,gpt_enhance_clip}, as shown in Figure \ref{fig:contra} (b), introduce \textbf{textual attribute} descriptions obtained from Large Language Models~\cite{gpt3,gpt4,llmsurvey}. These textual attribute descriptions are appended to the class name to construct text attribute prompts enriched with more semantics. The final classification result is determined by matching scores between the global image feature and the outputs of text attribute prompts across categories.

\begin{figure}[t]
  \centering
  \scalebox{0.9}{   \includegraphics[width=0.99\linewidth]{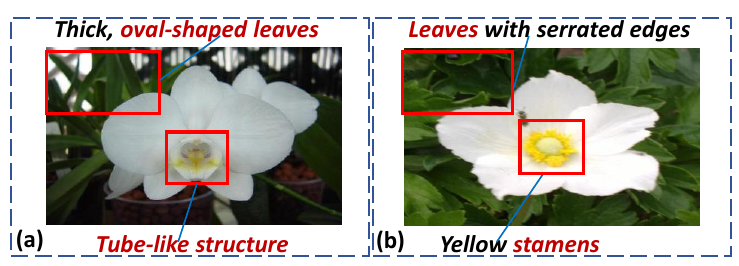}}

   \caption{(a) Moon Orchid and (b) Japanese Anemone exhibit strikingly similar overall appearances.  Visual attributes play a crucial role in distinguishing between them, such as the central yellow stamens of Japanese Anemone.}

   \label{fig:visual_attribute}
\end{figure}

Despite the performance improvements demonstrated by prior methods, two crucial aspects have been overlooked. \textbf{(1) Visual Attribute Modeling.} Previous methods rely on a single global image feature for classification (see Figure \ref{fig:contra} (a) and (b)). However, global image features may fall short in capturing fine-grained visual attribute information crucial for distinguishing visually similar classes in few-shot scenarios. As shown in Figure~\ref{fig:visual_attribute}, the Moon Orchid and Japanese Anemone exhibit quite similar overall appearances, making it challenging to differentiate between them relying solely on global features. However, distinguishing them becomes much easier by relying on their distinct leaf shapes and reproductive structures.  \textbf{(2) Attribute-Level Alignment.} 
The open-vocabulary visual recognition ability of the CLIP model stems from its global alignment between global image features and textual descriptions. However, when adapted to unseen tasks, the global alignment may lack robustness against disruptions from complex image backgrounds and irrelevant image details, hampering the image recognition ability. While previous methods have attempted to model class-specific textual attributes, as depicted in Figure \ref{fig:contra} (b), they still focus on alignment with the global image features and fall short in addressing disruptions present in images. To address this issue, in addition to the global alignment, establishing \textbf{attribute-level alignment} is imperative, \textit{i.e}., alignment between fine-grained visual and textual attribute features (see Figure \ref{fig:contra} (c)). This alignment empowers the model to selectively emphasize the distinctive visual attribute features described in the textual attributes, thereby enhancing the ability to handle disruptions in images.

Inspired by the above insights, we propose  \textbf{Multi-modal Attribute Prompting} (MAP) by jointly exploring textual attribute prompting, visual attribute prompting, and attribute-level alignment to enhance the adaptability of CLIP in downstream few-shot tasks.  For \textbf{textual attribute prompting}, we generate class-specific textual descriptions using a pre-trained large language model. Subsequently, these textual descriptions are utilized to create multiple textual attribute prompts, each encompassing context words, the class name, and an attribute description. It's challenging to directly capture appropriate discriminative visual attributes in an unknown test image without prior information. Hence, for \textbf{visual attribute prompting}, first, we use learnable initial visual attribute prompts to aggregate regional features by interacting with image tokens. Then, we utilize the specially designed \textbf{Adaptive Visual Attribute Enhancement} (AVAE) module, in which the initial visual attribute prompts are enhanced by adaptively selected textual attribute prompts. Through interaction with both image tokens and textual attribute prompts, visual attribute prompts can adaptively capture visual attribute features in an unseen image. Finally, we reformulate the \textbf{attribute-level alignment} between visual attribute prompts and textual attribute prompts as an Optimal Transport problem~\cite{OT} and use the Sinkhorn algorithm~\cite{sinkhorn} to solve it. The ultimate classification result is determined by both the global matching score and the attribute-level matching score. This integration of additional attribute alignment, alongside global alignment, achieves multi-level robust alignment between images and text categories.

Our main contributions can be summarized as follows:
\begin{itemize}
\item We propose \textbf{Multi-modal Attribute Prompting}, which jointly explores textual attribute prompting, visual attribute prompting, and attribute-level alignment between images and text categories. 
To our knowledge, this is the first work to model visual attributes and establish attribute-level alignment between images and text categories for adapting the pre-trained CLIP model to downstream few-shot tasks.
\item Extensive experimental results on 11 benchmark datasets demonstrate that our method performs favorably against state-of-the-art approaches.
\end{itemize}

\section{Related Works}\label{sec2}
In this section, we introduce several lines of research in pre-trained vision-language models and prompt learning.

\subsection{Vision-Language Models.} 

In recent years, pre-trained vision-language 
models~\cite{mei2022guest,zhang2020language, yu2019multimodal,yang2020grounding,flava, lit,Florence} have shown exceptional 
performance in diverse downstream tasks. Among them,  CLIP~\cite{CLIP} stands out as a representative approach. By training its vision and text encoders to map both modalities closely in a shared embedding space, CLIP establishes a comprehensive global alignment between images and their corresponding textual descriptions, enabling open-vocabulary classification tasks. The classification result can be obtained by computing the similarity scores of the global image feature with class names encoded by the text encoder. However, as classification relies solely on the global matching score, the accuracy may be affected by disruptions in images, such as complex backgrounds, especially in few-shot settings~\cite{jiang2020multi-metric-learning,cheng2021meta-learning,wang2023few,xu2022gct_few,zhang2022mfnet_few,zhang2019few,dang2023counterfactual_few}, where only a few training samples are available. To improve the robustness of cross-modal alignment, we achieve multi-level alignment for CLIP by introducing additional attribute-level alignment between dynamically learned textual and visual attribute features. In this manner, our method enhances the fine-grained perception capability with the pre-trained global knowledge preserved. 

\subsection{Prompt Learning.} 
Prompt learning is initially introduced in the field of natural language processing (NLP)~\cite{nlp_p3,nlp_p4_tune,nlp_p5_tune,nlp_p6_ppt,nlp_p7gpt}. With language models frozen, prompt learning methods effectively facilitate the adaptation of pre-trained language models to downstream few-shot tasks by involving additional hand-crafted or learnable prompt tokens. Prompt learning has recently been employed to enhance the adaptation of the CLIP model to downstream few-shot tasks, where limited training samples are available. CoOp~\cite{CoOp}  constructs prompts by concatenating learnable continuous vectors and class name tokens. CoCoOp~\cite{cocoop} extends CoOp by further learning a lightweight neural network to generate an input-conditional vector for each image, tackling the poor generalizability to broader unseen classes in CoOp~\cite{CoOp}. ProDA~\cite{ProDa} optimizes a set of prompts by learning the distribution of prompts. Instead of focusing on text-modal prompts, VPT~\cite{vpt} introduces learnable vectors to the Vision Transformer~\cite{vit} to refine image features within the frozen vision encoder. DAPT~\cite{dapt}, RPO~\cite{rpo}, and MaPLe~\cite{Maple} improve the generalization ability of VLMs via multimodal prompting. PromptSRC~\cite{PromptSRC} introduces regularization loss to prompt learning.
These methods rely solely on class names for text prompt construction and may struggle to fully encapsulate categorical semantics.

\subsection{Textual Attribute Prompts.} 
To enrich the semantic description for different classes,  recent works~\cite{classification_via_des,gpt_enhance_clip, Multi-des}, instead of relying solely on class names, have shifted towards the utilization of attribute descriptions to construct textual attribute prompts for each class. This shift is facilitated by the development of pre-trained large language models (LLMs) like the GPT family~\cite{gpt3,gpt4}. Attribute descriptions can be easily obtained by
querying the LLM with suitable question templates. However, these methods focus on attributes in text space only, neglecting the modeling of visual attributes, leading to limited visual perception capabilities of the model and misalignment between global visual and local textual features. In contrast, we jointly model visual and textual attribute features and establish attribute-level alignment between images and text categories.

\subsection{{Visual Attributes.}}
{
Visual attributes refer to intuitive properties of objects, encompassing low-level semantics (e.g., color, texture, and shape) and high-level semantics (e.g., head, body, and tail of objects)~\cite{Learning_visual_attributes_7}. Utilizing visual attributes has led to significant progress in various vision tasks, including image search~\cite{visual_attribute_search_17}, image recognition~\cite{visual_attribute_regconition_37}, and scene understanding~\cite{visual_attribute_scene_25}. Some previous works on learning attributes~\cite{visual_attribute_10,visual_attribute_search_17,visual_attribute_49}  usually require extensive manual attribute annotations, which are labor-intensive. Dealing with this issue, a recent work~\cite{visual_attribute_43} developed an encoder-decoder network to unsupervisedly distill high-level attribute-specific vectors without requiring attribute annotations. VAPNet~\cite{VAPNet} achieves semantic details by utilizing local image patches to distill visual attributes from these discovered semantics. Different from these methods, our approach uniquely leverages visual prompts to model visual attributes. By incorporating visual attribute prompts as learnable tokens within Vision Transformers, our method captures and aggregates relevant image features effectively.
}

\section{Methodology}\label{method}
\begin{figure*}[t]
  \centering
       \includegraphics[width=0.99\linewidth, ]{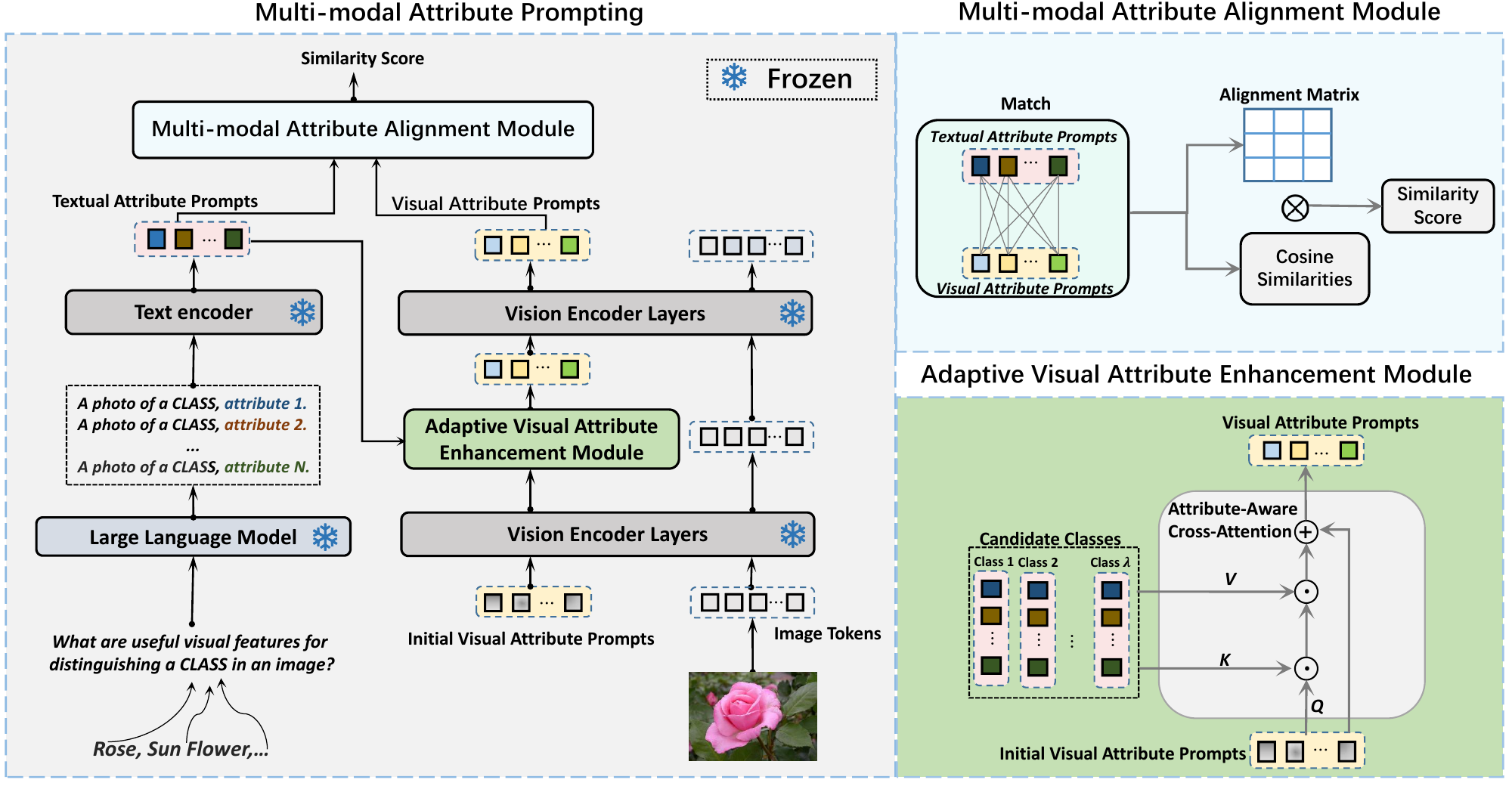}

  \caption{The architecture of our method: \textbf{MAP} leverages textual attribute descriptions to construct textual attribute prompts and incorporates learnable visual attribute prompts for capturing visual attributes. In the \textbf{Adaptive Visual Attribute Enhancement} module, initial visual attribute prompts are enhanced by textual attribute prompts via the attribute-aware cross-attention layer. The \textbf{Multi-modal Attribute Alignment} module calculates the similarity score between visual attributes and textual attributes with the optimal transport.}
  \label{fig:method}
\end{figure*}

In this section, we first provide a concise overview of CLIP~\cite{CLIP}. Then, we present a comprehensive introduction to our proposed multi-modal attribute prompting, as illustrated in Figure \ref{fig:method}, including textual attribute prompting, visual attribute prompting, and attribute-level alignment. The main symbols and instructions are shown in Table~\ref{tab:symbol-table}.

\begin{table}[]
    \centering
 
    \caption{Main symbols and instructions}
    \renewcommand\arraystretch{1.5} 
    \setlength{\tabcolsep}{2mm}{
    \begin{tabular}{c|c}
    \hline
    Symbol  &  Instruction \\ \hline
    $\phi(\cdot)$     & the image encoder \\ \hline
    $\theta(\cdot)$     & the text encoder \\ \hline
    $\mathcal{V}$  & set of class names   \\ \hline
    $\mathcal{C}$  & the number of class names   \\ \hline
    $x$  & the input image   \\ \hline
    $y$  & the ground-truth label    \\ \hline
    $f$  & the global image feature   \\ \hline

    $p_k^n$  &  the $n$-th textual attribute prompt of $k$-th class  \\ \hline
$g_k^n$  &  encoded $n$-th  textual attribute prompt of $k$-th class  \\ \hline
    
    $\boldsymbol{G}_k$  & encoded textual attribute prompts of the $k$-th class  \\ \hline
    $l_j$  & the $j$-th ViT layer    \\ \hline
    $E_j$ & image tokens output from $j$-th ViT layer \\ \hline
    $s_j$ & [CLS] token output from $j$-th ViT layer  \\ \hline
    $U_j$ & visual attribute prompts output from $j$-th ViT layer \\ \hline
    $\boldsymbol{F}$ & visual attribute prompts output from ViT \\ \hline
    $T^*$  & the optimal transportation plan  \\ \hline
    $\mathbf{\Gamma}$  &  adaptive visual attribute enhancement module  \\  \hline
$\psi(\cdot,\cdot)$  & similarity function \\ \hline
    
  { $M$}  & {the number of visual attribute prompts}  \\ \hline
    {$N$}  & {the number of textual attribute prompts } \\ \hline
      {$L$ } & {the number of transformer layers in ViT } \\ \hline
        {$\boldsymbol{Q}$, $\boldsymbol{K}$, $\boldsymbol{V}$ } &{queries, keys, and values in the attention layer } \\ \hline 
        {$W_Q$,$W_K$, $W_V$}    &{linear projections of the attention layer}  \\ \hline
        {$\mathbf{1}_N$ }   & {N-dimensional all-one vector } \\ \hline
       {$p$,$q$ }   & {discrete distributions}  \\ \hline
        {$\mu$,$\nu$}    & {discrete probability vectors}  \\ \hline

    \end{tabular}
    }
    \label{tab:symbol-table}
\end{table}

\subsection{Review of CLIP}\label{sec:overview}
The Contrastive Language-Image Pre-training (CLIP) model~\cite{CLIP} is a well-known vision-language model trained on large-scale image-text pairs. CLIP consists of two primary components: an image encoder $\phi(\cdot)$ for converting input images into visual embeddings and a text encoder $\theta(\cdot)$ for encoding textual information. During pre-training, CLIP trains encoders using a contrastive loss objective~\cite{contra}, with the purpose of achieving a global alignment between images and textual descriptions. The CLIP model can be easily applied to downstream tasks.
  
Given a set $\mathcal{V}$ of $\mathcal{C}$ class names, the text prompts $\{{t}_i \}_{i=1}^C$ are formulated as manually designed templates, such as `A photo of a [CLASS].' The classification vectors $\{w_i\}_{i=1}^C $ are derived by passing text prompts $\left\{{t}_i\right\}_{i=1}^C$ to the text encoder: $w_i=\theta(t_i)$. Given an image $x$ and its label $y$, the global image feature $f$ is extracted by the image encoder: $f=\phi(x)$. The classification probability is formulated as
\begin{equation}
  P(y=i|{x})=\frac{\exp \left(\cos \left({w}_i, {f}\right) / \tau\right)}{\sum_{j=1}^C \exp \left(\cos \left({w}_j, {f}\right) / \tau\right)},
  \label{eq:clip_p}
\end{equation}
where $\tau$  is a temperature parameter and $\cos (\cdot,\cdot)$  denotes the cosine similarity.

\subsection{Textual Attribute Prompting}\label{sec:group}
To address the potential `lexical weak tie' issue of relying solely on class names for text prompt construction, we create multiple textual attribute prompts for each class, which helps enrich the semantic content in text prompts.

\textbf{Attribute Descriptions.} Consistent with previous methods~\cite{gpt_enhance_clip,classification_via_des,Multi-des}, we obtain category attribute descriptions by querying a Large Language Model (LLM) using a predefined question template: `What are useful visual features for distinguishing a [CLASS] in an image?' In response, the LLM provides discriminative attribute descriptions for the queried class. We select  $N$ descriptions for each class from the query results.


\textbf{Textual Attribute Prompt Construction.} We formulate $N$ textual attribute prompts for each class by combining attribute description sentences with a standardized prompt template. For instance, for the $k$-th class, with the template `A photo of a [CLASS]' we construct a textual attribute prompt: $p_k^n$=\{A photo of a class ($k$), $t_k^n$\}, where class ($k$) denotes the class name corresponding to the $k$-th class, and $t_k^n$ denotes the $n$-th attribute description for the $k$-th class. To enhance the adaptability of textual attribute prompts, we replace the hand-crafted context, \textit{i.e}., `A photo of a' with several learnable context vectors. 
{Following CoOp~\cite{CoOp}, we use four learnable class-agnostic context vectors, concatenated with the class name and attribute description to construct the textual attribute prompt. These vectors are optimized during training to better adapt to downstream tasks, providing a more flexible context.}

By feeding the textual attribute prompts into the text encoder $\theta$, we can obtain encoded textual attribute prompts:
\begin{equation}
 \boldsymbol{G}_k=\{{g}_k^n|_{n=1} ^N\}, g_k^n=\theta(p_k^n),
  \label{eq:G_k}
\end{equation}
where $\boldsymbol{G}_k$ is the textual attribute prompt set for the $k$-class.

\subsection{Visual Attribute Prompting}
To improve fine-grained visual perception, we model visual attributes with visual attribute prompts. However, it is challenging to directly learn discriminative visual attributes for an unknown image without prior information. Therefore, we design an adaptive visual attribute enhancement module to adaptively establish visual attribute prompts under the guidance of textual attribute information.

\textbf{Learnable Visual Attribute Prompts.} We model visual attributes by introducing $M$ visual attribute prompts $U=\{ u_i\}_{i=1}^M$, where each attribute prompt $u_i$ is a randomly initialized learnable vector with the dimension of $d_v$. $\{ u_i\}_{i=1}^M$ are inserted into the first Vision Transformer (ViT) layer and are then propagated into deeper layers.
For the $j$-th ViT layer $l_j$, visual attribute prompts $U_{j-1}$ output from the ($j$-1)-th ViT layer are concatenated with image tokens $E_{j-1}$ and the learnable classification token $s_{j-1}$ ([CLS]), forming the input sequence of the current layer.
Formally,
\begin{equation}
[s_j,U_j,E_j]=l_j([s_{j-1},U_{j-1},E_{j-1}]), j=1,2,...,L,
  \label{eq:vit}
\end{equation}
where $[\cdot,\cdot]$ indicates the concatenation along the sequence length dimension. In early layers of ViT, the visual attribute prompts progressively aggregate image regional features through interaction with image tokens { facilitated by the attention mechanism. Learnable visual attribute prompts compute similarity with image tokens and aggregate information accordingly. Similar to the [CLS] token in models like BERT~\cite{2018bert} and ViT~\cite{vit}, visual prompts can read and aggregate visual information from image tokens~\cite{rpo}.
Previous research~\cite{ViT_see_like,fromcliptodino} indicates that ViTs will attend to local information in early layers. This property, together with the attention mechanism, helps aggregate image regional features.
}

\textbf{Adaptive Visual Attribute Enhancement Module.} AVAE, represented as $\mathbf{\Gamma}$, is designed to dynamically refine visual attribute prompts with textual attribute guidance for arbitrary images from unseen classes. As the category of the test image is unknown, we select possibly related textual attribute prompts from the most similar classes. Specifically, we first compute the similarities between the global image feature, \textit{i.e}., the classification token $s$, and textual category embeddings represented by the mean of textual attribute prompts.
Based on these similarities, we select the most similar $\lambda$ categories as the candidate classes 
and gather their textual attribute prompts as $\boldsymbol{G^{\prime}}=\{g_j|_{j=1}^{\lambda N}\}$. Subsequently, the textual attribute prompts 
 $\boldsymbol{G^{\prime}}$
are employed as the semantic guidance to enhance visual attribute prompts at the $l$-th ViT layer:
\begin{equation}
\{{\Tilde{u}_i}^{(l)}\}_{i=1}^M=\mathbf{\Gamma}( \{{u_i}^{(l)}\}_{i=1}^M,\boldsymbol{G^\prime}),
  \label{eq:gamma}
\end{equation}
where $\mathbf{\Gamma}$ takes the  initial visual attribute prompts $\{ {u_i}^{(l)}\}_{i=1}^M$ generated from 
$l$-th layer as the input, and refine them conditioned on textual attribute prompts   $\boldsymbol{G^\prime}$. 
Then the enhanced visual attribute prompt ${\Tilde{u}_i}^{(l)}$ is inserted into the ($l+1$)-th layer for progressive attribute learning.

To better inject the semantic clues of selected textual prompts into visual attribute prompts, 
we design an attribute-aware cross-attention layer in $\mathbf{\Gamma}$. Here, the visual attribute prompt tokens $\{{u_i}^{(l)}\}_{i=1}^M$ function as queries $\boldsymbol{Q}$. Simultaneously, the textual attribute prompt features $\boldsymbol{G^{\prime}}$ of candidate classes are utilized as keys $\boldsymbol{K}$ and values $\boldsymbol{V}$. The enhanced visual attribute prompt $\Tilde{u}_i^{(l)}$ is formulated as 

\begin{gather}
\Tilde{\alpha}_{i j}=\frac{\exp({\alpha_{ij}})}{\sum_{j^\prime=1}^{\lambda N} \exp({\alpha_{ij^\prime}})},
{\alpha}_{i j}=\frac{u_i^{(l)}W_Q\cdot (g_jW_K)^T}{\sqrt{d_K}},\\
\Tilde{u}_i^{(l)}=u_i^{(l)}+\sum_{j=1}^{\lambda N }\Tilde{\alpha}_{i j} (g_jW_V), i=1,2, \cdots, \lambda N,
  \label{eq:u_n}
\end{gather}
where $W_Q$,$W_K$ and $W_V$ are linear projections of the attention layer. {Attention scores} $\Tilde{\alpha}_{i j}$ indicate the correspondence between visual and textual attribute prompts, emphasizing relevant image-specific semantic attribute patterns for enhancing the visual attribute prompts. After the text-guided enhancement, the {refined visual attribute prompts $\{\Tilde{u}_i^{(l)}\}_{i=1}^M$ are propagated into the remaining vision encoder layers and continue to capture visual attributes through interaction with image tokens.}

\subsection{Attribute-Level Alignment}
To achieve precise alignment between visual attribute prompts $ \{{u_i}^{(L)}\}_{i=1}^M$ and textual attribute prompts $\boldsymbol{G_k}=\{g_k^n|_{n=1}^N\}$, we formulate the attribute-level matching task as an Optimal Transport (OT) problem~\cite{OT}. For simplicity, we refer to $ \{{u_i}^{(L)}\}_{i=1}^M$ as $\boldsymbol{F}=\{f_m|_{m=1}^M\} $ hereafter. 
{Optimal Transport (OT)~\cite{OT} is a powerful tool to measure the distance between two distributions. 
Given two sets of feature points $\boldsymbol{F}=\{f_m|_{m=1}^M\} $ and $\boldsymbol{G_k}=\{g_k^n|_{n=1}^N\}$, their distributions can be formulated as $p=\sum_{m=1}^M \mu_m \delta _{f_m}$, $q=\sum_{n=1}^N \nu_n \delta_{g_k^n}$, $\delta _{f_m} $ is a Dirac delta function centered at a specific point $f_m$ in the embedding space.
Here, $\mu \in \mathbb{R}^M$, $\nu \in \mathbb{R}^N$ are two discrete distribution vectors. We define the cost matrix between $\boldsymbol{F}=\{f_m|_{m=1}^M\} $ and $\boldsymbol{G_k}=\{g_k^n|_{n=1}^N\}$ as $\mathbf{C} \in \mathbb{R}^{M \times N}$, where $\mathbf{C}_{m,n}=1-\langle f_m,g_k^n \rangle$ is the transport cost from $f_m$ to $g_k^n$. The transport cost between $p$ and $q$ is $\langle 
 \mathbf{T},\mathbf{C} \rangle$, where $\mathbf{T}$ is the transport plan,
and $\mathbf{T}_{m,n}$ is the probability or ``flow" of transporting from $f_m$ to $g_k^n$. The goal of OT is to transport $p$ to $q$ at the smallest cost with the optimal transport plan $\mathbf{T}^*$:
\begin{equation}
\begin{gathered}
\mathbf{T}^*=\underset{\mathbf{T} \in \Pi(p, q)}{\arg \min }\langle\mathbf{T}, \mathbf{C}\rangle, \\
\text { s.t. } \quad \mathbf{T} \mathbf{1}_N=\mu, \mathbf{T}^T \mathbf{1}_M=\nu,
\end{gathered}
  \label{eq:ot_cost}
\end{equation}
where $\prod(p, q)$ is the joint distribution with marginals $\mu$ and $\nu$, and $\langle\cdot,  \cdot\rangle$ denotes the Frobenius inner product. To accelerate the solving process, we use the Sinkhorn algorithm, which introduces the entropic regularization term to the transport cost to encourage smoother solutions: $\underset{\mathbf{T}}{\min }\langle\mathbf{T}, \mathbf{C}\rangle -\gamma h(\mathbf{T})$, $\gamma$ is a constant hyperparameter controlling the intensity of regularization term.
Instead of solving the constrained optimization directly, the Sinkhorn algorithm~\cite{sinkhorn} employs an iterative procedure:
\begin{equation}
\begin{gathered}
\mathbf{T}^*=diag(U(t))\mathbf{A}diag(V(t)), \\
\mathbf{A}=exp(-\mathbf{C}/\gamma)
\end{gathered}
  \label{eq:ot_cost2}
\end{equation}
where in the $t$-th iteration, $U(t)=\mu/(\mathbf{A}V(t-1))$, $V(t)=\nu/\mathbf{A}^TU(t))$, with the initiation $V(0)=\mathbf{1}$.
With Equation ~\eqref{eq:ot_cost2},} we can obtain $\mathbf{T}^*$ to serve as the alignment matrix, and then define the final similarity score between the visual attribute prompts $\boldsymbol{F}$ and textual attribute prompts $\boldsymbol{G}_k$ as:
\begin{equation}
\psi(\boldsymbol{F},\boldsymbol{G}_k)=\sum_{m=1}^M \sum_{n=1}^N \langle f_m,g_k^n \rangle \mathbf{T}_{m,n}^*,
  \label{eq:sim}
\end{equation}
where $\psi(\cdot,\cdot)$ denotes the similarity function.

\subsection{Training Objectives}

Based on the attribute-level alignment, we can classify the image $x$ with fine-grained visual attributes:
\begin{equation}
\begin{gathered}
P_a(y=i| {x})=\frac{\exp( \psi \left(\left(\boldsymbol{F}, \boldsymbol{G}_i\right) / \tau\right))}{\sum_{j=1}^C \exp (\psi (\boldsymbol{F}, \boldsymbol{G}_j / \tau))}.
\end{gathered}
  \label{eq:P_a}
\end{equation}
Furthermore, relying on the global alignment in CLIP, the prediction probability is computed as
\begin{equation}
\begin{gathered}
P_g(y=i| {x})=\frac{\exp( \cos\left(\left(\boldsymbol{f}, \overline{\boldsymbol{g}}_i\right) / \tau\right))}{\sum_{j=1}^C \exp (\cos (\boldsymbol{f}, \overline{\boldsymbol{g}}_j / \tau))},
\end{gathered}
  \label{eq:P_g}
\end{equation}
where $\boldsymbol{f}$ is the global feature of the image $x$, \textit{i.e}., the class token $s_L$, and $\overline{\boldsymbol{g}}_i$ is the textual categorical embedding of the $i$-th class, \textit{i.e}., the mean value of textual prompts in $\boldsymbol{G}_i$. The final prediction probability is 
\begin{equation}
\begin{gathered}
P(y=i| {x})=P_g(y=i| {x})+\beta P_a(y=i|{x}),
\end{gathered}
  \label{eq:P_f}
\end{equation}
which incorporates both global-level prediction scores and additional attribute-level matching scores, achieving multi-level robust alignment between images and categorical texts. Naturally, the classification loss is formulated as:
\begin{equation}
\begin{aligned}
 L_{cls}=-\frac{1}{B}\sum_{i=1}^B log(P(y=y_i|x_i)),
\end{aligned}
  \label{eq:t_loss}
\end{equation}
where $B$ is the batch size of image-text pairs, and $y_i$ denotes the ground-truth label  of  the input image $x_i$.

\section{Experiments}

\begin{table*}[t]
  \caption{{Comparison with CLIP, CoOp and CoCoOp in the base-to-novel generalization setting.}  The results demonstrate the strong generalizability to novel classes of our MAP. HM: Harmonic mean to highlight the generalization trade-off~\cite{HM}. The best results in each column are shown in \textbf{bold} font.}
 
  \begin{minipage}[t]{0.3\linewidth}
    \centering
    
    \caption*{   \centering  (a) \textbf{Average results }} 
    \scalebox{1.2}{

  \begin{tabular}{@{\hspace{4pt}}c@{\hspace{4pt}}c@{\hspace{6pt}}c|c@{}}
      \toprule
        Method & Base &  {Novel} & HM \\
      \midrule
      CLIP & 69.34 & 74.22 & 71.70 \\
      CoOp & 82.69 & 63.22 & 71.66 \\
      CoCoOp & 80.47 & 71.69 & 75.83 \\
      \rowcolor[HTML]{EFEFEF}
       Ours & \textbf{83.66} & \textbf{75.76} & \textbf{79.36} \\
      \bottomrule
    \end{tabular}
    }
    \label{fig:avg}
    \vspace{+0.4cm}

      \caption*{  \centering  (d) DTD} 
\scalebox{1.2}{
  \begin{tabular}{@{\hspace{4pt}}c@{\hspace{4pt}}c@{\hspace{6pt}}c|c@{}}
        \toprule
        Method  & Base &  {Novel} & HM \\
        \midrule
        CLIP & 53.24 & 59.90 & 56.37 \\
        CoOp & 79.44 & 41.18 & 54.24 \\
        CoCoOp & 77.01 & 56.00 & 64.85 \\
        \rowcolor[HTML]{EFEFEF}
         Ours & \textbf{82.63} & \textbf{66.23} & \textbf{73.53} \\
      
        \bottomrule
      \end{tabular}
      }
      \label{fig:dtd}
\vspace{+0.4cm}

    \caption*{    \centering (g) OxfordPets} 
  \scalebox{1.2}{

  \begin{tabular}{@{\hspace{4pt}}c@{\hspace{4pt}}c@{\hspace{6pt}}c|c@{}}
      \toprule
       Method & Base &  {Novel} & HM \\
      \midrule
      CLIP & 91.17 & 97.26 & 94.12 \\
      CoOp & 93.67 & 95.29 & 94.47 \\
      CoCoOp & 95.20 & \textbf{97.69} & \textbf{96.43} \\
      \rowcolor[HTML]{EFEFEF}
       Ours & \textbf{95.43} & 96.90 & 96.16 \\     
      \bottomrule
    \end{tabular}
    }
    \label{fig:pets}
     \vspace{+0.4cm}
    \caption*{   \centering (j) Food101} 
 \scalebox{1.2}{

  \begin{tabular}{@{\hspace{4pt}}c@{\hspace{4pt}}c@{\hspace{6pt}}c|c@{}}
      \toprule
       Method & Base &  {Novel} & HM \\
      \midrule
      CLIP & 90.10 & 91.22 & 90.66 \\
      CoOp & 88.33 & 82.26 & 85.19 \\
      CoCoOp & \textbf{90.70} & \textbf{91.29} & \textbf{90.99} \\
      \rowcolor[HTML]{EFEFEF}
       Ours & 90.30 & 89.30 & 89.80 \\
    
      \bottomrule
    \end{tabular}
    }
    \label{fig:food}
     \vspace{+0.4cm}
   
    \label{fig:subfigure-a}
  \end{minipage}
  \hspace{0.6cm}
  \begin{minipage}[t]{0.3\linewidth}
    \centering
   
     \caption*{   \centering (b) ImageNet} 
\scalebox{1.2}{

  \begin{tabular}{@{\hspace{4pt}}c@{\hspace{4pt}}c@{\hspace{6pt}}c|c@{}}
      \toprule
       Method & Base &  {Novel} & HM \\
      \midrule
      CLIP & 72.43 & 68.14 & 70.22 \\
      CoOp & 76.47 & 67.88 & 71.92 \\
      CoCoOp & 75.98 & 70.43 & 73.10 \\
      \rowcolor[HTML]{EFEFEF}
       Ours & \textbf{76.60} & \textbf{70.60} & \textbf{73.48} \\
        
      \bottomrule
    \end{tabular}
    }
    \label{fig:imagenet}
    \vspace{+0.4cm}

   \caption*{   \centering (e) EuroSAT} 
\scalebox{1.2}{
  \begin{tabular}{@{\hspace{4pt}}c@{\hspace{4pt}}c@{\hspace{6pt}}c|c@{}}
        \toprule
        Method  & Base &  {Novel} & HM \\
        \midrule
        CLIP & 56.48 & 64.05 & 60.03 \\
        CoOp & \textbf{92.19} & 54.74 & 68.69 \\
        CoCoOp & 87.49 & 60.04 & 71.21 \\
        \rowcolor[HTML]{EFEFEF}
         Ours & 92.13 & \textbf{76.10} & \textbf{83.33} \\     
        \bottomrule
      \end{tabular}

      }
      \label{fig:eurosat}
\vspace{+0.4cm}

    \caption*{   \centering (h) StanfordCars} 
  \scalebox{1.2}{
  
  \begin{tabular}{@{\hspace{4pt}}c@{\hspace{4pt}}c@{\hspace{6pt}}c|c@{}}
      \toprule
       Method & Base &  {Novel} & HM \\
      \midrule
      CLIP & 63.37 & \textbf{74.89} & 68.65 \\
      CoOp & \textbf{78.12} & 60.40 & 68.13 \\
      CoCoOp & 70.49 & 73.59 & 72.01 \\
      \rowcolor[HTML]{EFEFEF}
       Ours & 76.70 & 73.73 & \textbf{75.18} \\
   
      \bottomrule
    \end{tabular}
    }
    \label{fig:stanford}
     \vspace{+0.4cm}
    \caption*{   \centering (k) FGVCAircraft} 
  \scalebox{1.2}{
  
  \begin{tabular}
  {@{\hspace{4pt}}c@{\hspace{4pt}}c@{\hspace{6pt}}c|c@{}}
      \toprule
       Method & Base &  {Novel} & HM \\
      \midrule
      CLIP & 27.19 & 36.29 & 31.09 \\
      CoOp & 40.44 & 22.30 & 28.75 \\
      CoCoOp & 33.41 & 23.71 & 27.74 \\
      \rowcolor[HTML]{EFEFEF}
       Ours & \textbf{41.63} & \textbf{36.43} & \textbf{38.84} \\     
      \bottomrule
    \end{tabular}
    }
    \label{fig:fgvc}
     \vspace{+0.4cm}

  \end{minipage}
  \hspace{0.6cm}
  \begin{minipage}[t]{0.3\linewidth}
    \centering

    \caption*{   \centering    (c) Caltech101} 
 \scalebox{1.2}{
   \begin{tabular}{@{\hspace{4pt}}c@{\hspace{4pt}}c@{\hspace{6pt}}c|c@{}}
      \toprule
       Method & Base &  {Novel} & HM \\
      \midrule
      CLIP & 96.84 & 94.00 & 95.40 \\
      CoOp & 98.00 & 89.81 & 93.73 \\
      CoCoOp & 97.96 & \textbf{93.81} & 95.84 \\
      \rowcolor[HTML]{EFEFEF}
       Ours & \textbf{98.30} & 93.80 & \textbf{96.00} \\
    
      \bottomrule
    \end{tabular}
    }
    \label{fig:caltech101}
     \vspace{+0.4cm}

\caption*{   \centering (f) UCF101} 
  \scalebox{1.2}{
  
  \begin{tabular}{@{\hspace{4pt}}c@{\hspace{4pt}}c@{\hspace{6pt}}c|c@{}}
      \toprule
      Method  & Base &  {Novel} & HM \\
      \midrule
      CLIP & 70.53 & 77.50 & 73.85 \\
      CoOp & 84.69 & 56.05 & 67.46 \\
      CoCoOp & 82.33 & 73.45 & 77.64 \\
      \rowcolor[HTML]{EFEFEF}
       Ours & \textbf{86.67} & \textbf{78.77} & \textbf{82.52} \\   
      \bottomrule
    \end{tabular}
    }
    \label{fig:ucf101}

    \vspace{+0.4cm}

    \caption*{   \centering (i) Flowers102} 
\scalebox{1.2}{

  \begin{tabular}{@{\hspace{4pt}}c@{\hspace{4pt}}c@{\hspace{6pt}}c|c@{}}
      \toprule
       Method & Base &  {Novel} & HM \\
      \midrule
      CLIP & 72.08 & \textbf{77.80} & 74.83 \\
      CoOp & \textbf{97.60} & 59.67 & 74.06 \\
      CoCoOp & 94.87 & 71.75 & 81.71 \\
      \rowcolor[HTML]{EFEFEF}
       Ours & 97.57 & 75.23 & \textbf{84.95} \\

      \bottomrule
    \end{tabular}
    }
    \label{fig:flowers102}
 \vspace{+0.4cm}
    \caption*{   \centering (l) SUN397} 
  \scalebox{1.2}{
  
  \begin{tabular}{@{\hspace{4pt}}c@{\hspace{4pt}}c@{\hspace{6pt}}c|c@{}}
      \toprule
       Method & Base &  {Novel} & HM \\
      \midrule
      CLIP & 69.36 & 75.35 & 72.23 \\
      CoOp & 80.60 & 65.89 & 72.51 \\
      CoCoOp & 79.74 & \textbf{76.86} & 78.27 \\
      \rowcolor[HTML]{EFEFEF}
       Ours & \textbf{82.33} & 76.30 & \textbf{79.20} \\      
      \bottomrule
    \end{tabular}
    }
    \label{fig:sun397}
\vspace{+0.4cm}
 \end{minipage}
  \hspace{0.6cm}

  \label{tab:base-to-new}
\end{table*}

In this section, we begin by introducing the benchmark settings and implementation details, followed by a comprehensive presentation of the experimental results. 

All the models used are based on the open-source CLIP~\cite{CLIP} model. We evaluate the adaptation and generalization capability of MAP in four distinct settings following previous methdos~\cite{cocoop,CoOp}.

 \textbf{Base-to-novel generalization.}
 Datasets are split into base and novel classes. The model is trained on the training dataset, which is constructed by randomly selecting 16 images per class from base classes. Then the model is evaluated on both base and novel classes. The evaluation encompasses 11 image recognition datasets, including Food101 (Foo)~\cite{food101}, DTD~\cite{DTD}, ImageNet (Img)~\cite{imagenet}, Caltech101 (Cal)~\cite{cal}, EuroSAT (Eur)~\cite{eurosat}, StanfordCars (Car)~\cite{cars}, FGVCAircraft (FGV)~\cite{aircraft}, Flowers102 (Flo)~\cite{flowers}, OxfordPets (Pet)~\cite{pets}, 
 UCF101 (UCF)~\cite{pets}, and SUN397 (SUN)~\cite{sun}.

\textbf{Few-shot image classification.} To evaluate the learning capacity under extremely limited supervision, we assess the model's performance across varying shot scenarios, namely, 1, 2, 4, 8, and 16 shots. Similar to the base-to-novel generalization setting, we employ the same 11 datasets.

\textbf{Domain generalization.} To assess the robustness under domain shifts, we train the model using the source dataset ImageNet and subsequently evaluate its performance on out-of-distribution target datasets, namely ImageNet-R (-R)~\cite{imagenet-r}, ImageNet-A (-A)~\cite{imagenet-a}, ImageNetV2 (V2)~\cite{imagenetV2}, and ImageNet-Sketch (-S)~\cite{imagenet-s}.

\textbf{Cross-dataset evaluation.} In the cross-dataset transfer setting, we train the models on the source dataset ImageNet and directly evaluate them on target datasets. Specifically, the target datasets include Food101, DTD, Caltech101, EuroSAT, StanfordCars, FGVCAircraft, Flowers102, OxfordPets, UCF101, and SUN397.


\textbf{Implementation Details.} In all the experiments, we use the pre-trained CLIP ~\cite{CLIP} with ViT-B/16 image encoder backbone as the base model. We use the GPT-3.5 as the large language model. For MAP, we set the number of textual attribute prompts $N$ to 4, and the number of visual attribute prompts $M$ to 4. The AVAE module is inserted into the 7th transformer layer in the Vision Transformer (ViT). The default value of $\lambda$ is set as 10. $\beta$ is set as 1. We train the model using the SGD optimizer with a learning rate of 0.002. For the base-to-novel generalization setting, the model is trained for 20 epochs with a batch size of 16. For few-shot image classification, the maximum epoch is set to 200 for 16/8 shots, 100 for 4/2 shots, and 50 for 1 shot (except for ImageNet, where the maximum epoch is fixed to 50).


\subsection{Base-to-Novel Generalization}

\begin{table}
  \caption{Comparing MAP against more methods on the average accuracy over 11 datasets.}
   
  \centering
    \footnotesize
 \scalebox{1.3}{

  \begin{tabular}{ccc|c}
      \toprule
       Method & Base & Novel & HM \\
      \midrule
      CLIP~\cite{CLIP}  & 69.34 & 74.22 & 71.70 \\
      CoOp~\cite{CoOp} & 82.69 & 63.22 & 71.66 \\
      CoCoOp~\cite{cocoop} & 80.47 & 71.69 & 75.83 \\
      ProDA~\cite{ProDa}    & 81.56 & 72.30  & 76.65 \\
      RPO~\cite{rpo} & 81.13 & 75.00 & 77.78 \\      
      VDT-Adapter~\cite{gpt_enhance_clip} & 82.48 & 74.51 & 78.09 \\
      MaPLe~\cite{Maple} & 82.28 & 75.14 & 78.55 \\      
\rowcolor[HTML]{EFEFEF}
       MAP & \textbf{83.66} & \textbf{75.76} & \textbf{79.36} \\   
      \bottomrule
    \end{tabular}
}
  \label{tab:more_results}
\end{table}

                                 


To demonstrate generalization to label-shift, where labels are divided into base and novel classes for each dataset, we train the model on training datasets constructed by randomly selecting 16 images per class from base classes. The model is trained using this few-shot sampled data for 3 random seeds, and the results are averaged. We evaluate accuracy on test data corresponding to both the base and novel classes and use their harmonic mean~\cite{HM} as the final evaluation metric.

Compared to CoOp, MAP exhibits higher harmonic mean accuracy across all datasets. As shown in Table \ref{tab:base-to-new}, MAP, on average, increases novel accuracy by 12.54\% and base accuracy by 0.97\%. This demonstrates that MAP not only enhances the model's generalization to novel classes but also achieves better alignment between visual and textual modalities within base classes.

Compared to CoCoOp, MAP demonstrates superior generalization to novel classes, achieving an impressive average gain of up to 4.07\%. When considering both base and novel classes, MAP outperforms CoCoOp with an absolute average gain of 3.53\%. Among the 11 datasets, MAP exhibits higher accuracy than CoCoOp in 10 base datasets and 7 novel datasets. 

%

We present the average accuracy results across  11 datasets for MAP compared with several other methods in Table \ref{tab:more_results}. MAP outperforms other methods by a significant margin, demonstrating our superior performance over other methods. It's worth noting that VDT-Adapter~\cite{gpt_enhance_clip}, which leverages textual attributes obtained from GPT-4 to formulate prompts, improves the novel accuracy compared to CoOp. However, it neglects modeling visual attributes and fails to leverage the role of attributes fully. MAP outperforms VDT-Adapter 1.18\% in base classes and 1.25\% in novel classes.

\begin{figure*}[t]
  \centering
       \includegraphics[width=0.99\linewidth]{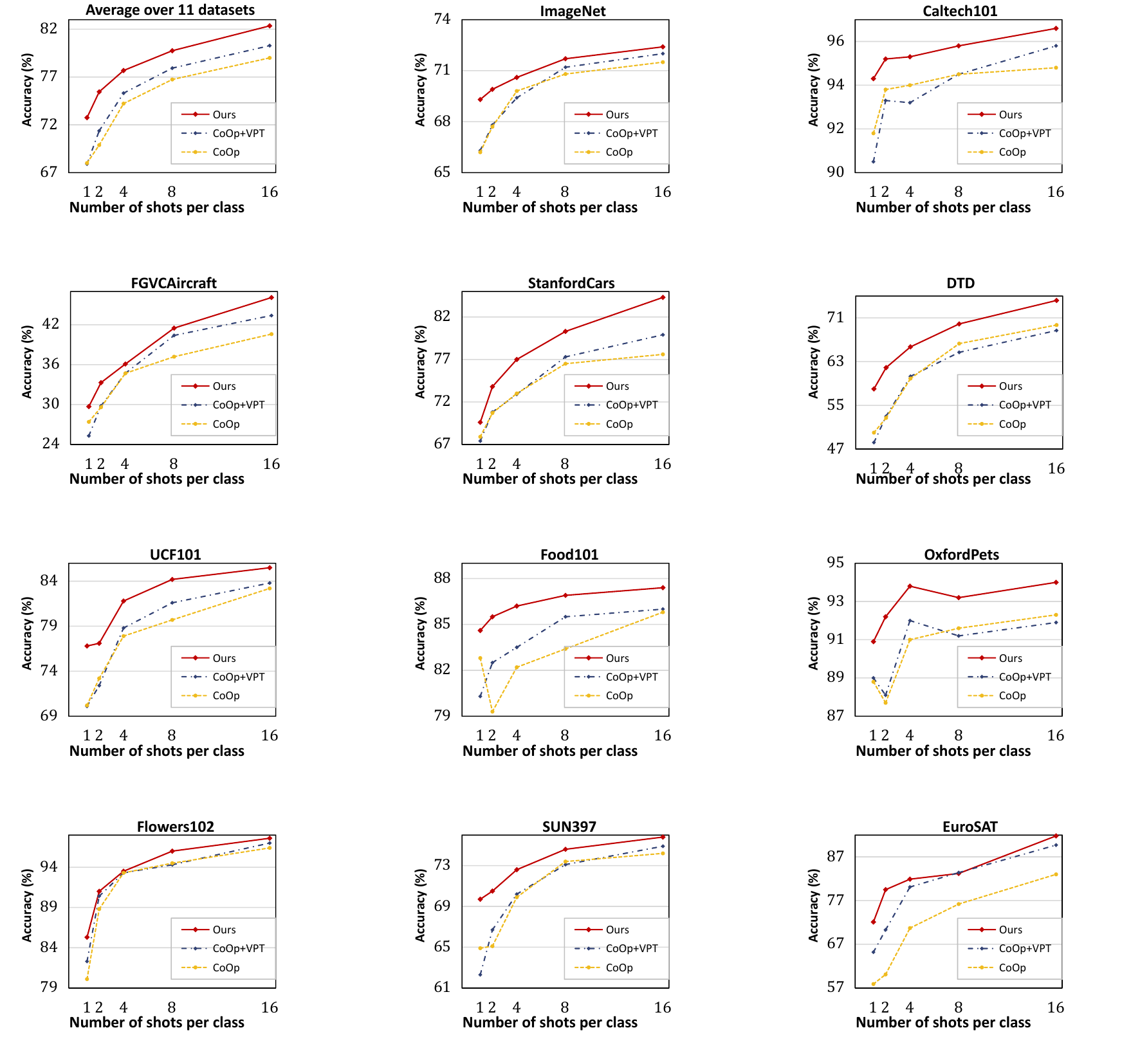}
  \hfill
  
\caption{{Main results of few-shot image classification on 11 datasets.
MAP consistently outperforms other CLIP adaptation methods across all datasets, demonstrating the strong few-shot adaptability of MAP.}}

  \label{fig:few-shot}
\end{figure*}

 \begin{figure}[]
 
  \centering
  \scalebox{0.8}{ \includegraphics[width=1\linewidth]{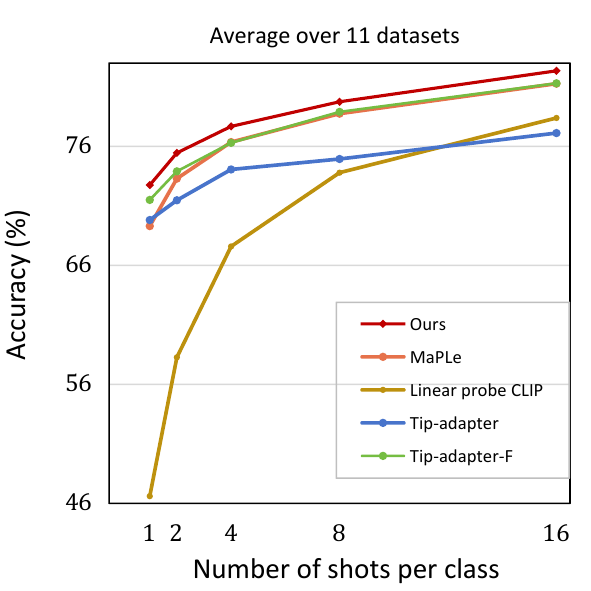}}

   \caption{{The average few-shot image classification results of more methods across 11 datasets. }}

   \label{fig:fs-more}
\end{figure}

\subsection{Few-Shot Image Classification}

To evaluate few-shot learning ability, we adopt the few-shot evaluation protocol from CLIP~\cite{CLIP}, utilizing 1, 2, 4, 8, and 16 shots per class for training and deploying models in full test sets. Figure \ref{fig:few-shot} summarizes the performance of MAP in few-shot learning on 11 datasets. Each plot compares MAP with CoOp and CoOp+VPT. CoOp+VPT refers to the combination of CoOp and VPT, \textit{i.e}., the integration of both learnable text prompts and learnable visual prompts~\cite{vpt} into the CLIP model simultaneously. In terms of the overall performance (Figure \ref{fig:few-shot}, top-left), compared to CoOp, the combination of CoOp and VPT  shows some improvement, though not significant.  However, in the 1-shot setting, the performance of the combination is even worse than CoOp alone. This suggests that simply introducing more learnable parameters in the vision encoder brings limited performance improvement in the extreme few-shot setting.
However, MAP consistently delivers significant performance improvements, even in scenarios with very few training samples (\textit{e.g}., 1-shot), showcasing the effectiveness of our visual attribute prompts enhanced by textual guidance. Furthermore, on certain datasets (Caltech101, Flowers102, DTD, SUN397, and OxfordPets), CoOp+VPT does not outperform CoOp alone, whereas MAP consistently achieves superior performance across all benchmark datasets, demonstrating the generalizability of MAP across diverse datasets.

In Figure \ref{fig:fs-more}, we present the performance results of additional methods for few-shot image classification. 
{
Tip-adapter-F~\cite{Tip-adapter}, the fine-tuned version of Tip-adapter, requires fine-tuning on the few-shot training data to update the adapter. The results show that Tip-adapter-F consistently achieves better performance than Tip-adapter and Linear probe CLIP. MaPLe~\cite{Maple} achieves performance comparable to Tip-adapter-F overall. Notably, MAP consistently outperforms both MaPLe~\cite{Maple} and Tip-adapter-F~\cite{Tip-adapter} in few-shot image classification across various shot settings, highlighting the effectiveness of our proposed approach.
}

\subsection{Domain Generalization }

To evaluate the model's robustness under domain shifts, we initially train the model using the source dataset, ImageNet~\cite{imagenet}. Subsequently, we evaluate its performance on target out-of-distribution datasets, namely ImageNetV2~\cite{imagenetV2}, ImageNet-Sketch~\cite{imagenet-s}, ImageNet-A~\cite{imagenet-a} and ImageNet-R~\cite{imagenet-r}. The overall results are summarized in Table \ref{tab:robustness}. From the experimental results, 
{the fully fine-tuned CLIP model shows poorer performance compared to the zero-shot CLIP on the ImageNet dataset and variants of ImageNet. This demonstrates that naive fine-tuning of the entire CLIP model may cause overfitting on the training set, leading to performance degradation.}
MAP achieves remarkable performance on unseen data compared to zero-shot CLIP~\cite{CLIP}, linear probe CLIP, CoOp~\cite{CoOp} and CoCoOp~\cite{cocoop}.
Compared to MaPLe, MAP shows slightly lower performance on ImageNet-Sketch but outperforms MaPLe~\cite{Maple} on other target datasets (ImageNetV2, ImageNet-A, and ImageNet-R). This underscores the robustness of MAP to domain shifts.


\subsection{Cross-Dataset Evaluation }

To demonstrate the model's capacity for generalization beyond a single dataset, we conduct training on ImageNet~\cite{imagenet} and subsequently evaluate its performance on the other 10 datasets.
{When transferring to other datasets, textual attribute prompts are constructed using class attribute descriptions of the target dataset classes, which are also collected from the LLM.  The learned parameters can be directly transferred, allowing effective inference despite category differences between the source and target datasets.
}
 Table \ref{tab:cross-data} presents a comprehensive overview of the performance comparison between MAP and previous methodologies on the cross-dataset evaluation benchmark.  On the source dataset, MAP achieves the highest score, underscoring its effectiveness in the source domain. When compared with CoOp~\cite{CoOp}, CoCoOp~\cite{cocoop}, and MaPLe~\cite{Maple}, MAP demonstrates a superior capacity for generalization across diverse datasets. Specifically, it outperforms these methodologies in 7 out of 10, 6 out of 10, and 6 out of 10 datasets, respectively. This suggests that MAP exhibits robustness to varied data distributions.

\begin{table*}[!tb]

    \small \centering
 \setlength{\tabcolsep}{8pt}
  \caption{{ {Domain generalization evaluation.} } Methods are trained on the source dataset ImageNet and evaluated on datasets with domain shifts, including ImageNetV2, ImageNet-S, ImageNet-A, and ImageNet-R.} 
  \vspace{+0.3cm}
    \scalebox{1.}[1.]{
    \begin{tabular}{l c@{\hspace{20pt}}ccccc}
    \toprule
    &  {Source} & \multicolumn{5}{c}{ {Target}} \\ \cmidrule(lr){2-2} \cmidrule(lr){3-7}
     & ImageNet & ImageNetV2 & ImageNet-S & ImageNet-A & ImageNet-R  & Avg.\\
    \midrule 
    CLIP~\cite{CLIP} &   {66.73} &  {{60.83}} & 46.15  & 47.77  & 73.96  & {57.18} \\
     {Fully Fine-Tuned CLIP} &   {61.65} &  {{52.70}} & {26.10}  & {17.55}  & {50.15}  & {36.63} \\

   Linear probe CLIP~\cite{CLIP} &   {67.42} &  {{57.19}} & 35.97  & 36.19  & 60.10  & {47.36} \\

    CoOp~\cite{CoOp} &   {71.51} &  {{64.20}} & 47.99  & 49.71  & 75.21  & {59.28} \\
    CoCoOp~\cite{cocoop} & 71.02 & {64.07} & 48.75 & 50.63 & 76.18 & {59.91}  \\
        MaPLe~\cite{Maple} & 70.72  & {64.07} & \textbf{49.15}  & 50.90 & 76.98 & {60.27}  \\
    
        MAP & \textbf{71.60}  & \textbf{64.47} & 49.07  & \textbf{51.07} & \textbf{77.37} & \textbf{60.49}  \\
             \midrule

    \end{tabular}
}
       
    \label{tab:robustness}
   
\end{table*}
  
\begin{table*}[!tb]

    \small \centering
 \setlength{\tabcolsep}{8pt}
  \caption{{ Cross-dataset evaluation.} Models are trained on ImageNet and evaluated on target datasets. MAP achieves overall favorable performance.} 

    \scalebox{1}[1]{
    \begin{tabular}{l cccccccccccc}
    \toprule
    &  {Source} & \multicolumn{10}{c}{ {Target}} \\ \cmidrule(lr){2-2} \cmidrule(lr){3-12}
     & ImageNet  &  {Cal} &  {Pet} &  {Car} &  {Flo} &  {Foo} &  {Air} &  {SUN} &  {DTD} &  {Eur} &  {UCF}\\
    \midrule 
    CoOp~\cite{CoOp} &   {71.51} & 93.70 & 89.14 & 64.51 & 68.71 & 85.30 & 18.47 & 64.15 & 41.92 & {{46.39}} & 66.55  \\
    CoCoOp~\cite{cocoop} & 71.02 &  \textbf{ 94.43} & {90.14} & 65.32 & {71.88} & 86.06 & 22.94 &   {67.36} & 45.73 & 45.37 & 68.21  \\
    MaPLe~\cite{Maple} & 70.72 & 93.53 &   {{90.49}} & \textbf{65.57} &  \textbf{ {72.23}} &   \textbf{{86.20}} &   {{24.74}} & 67.01 & {46.49} & \textbf{  {{48.06}}} & {68.69} &    \\
    \midrule
MAP & \textbf{71.60} &   {93.93} & \textbf{{90.80}} &   {63.00} & {68.40} & {{86.07}} & \textbf{{24.87}} & \textbf{68.10} &   \textbf{51.87} & {{42.63}} &   \textbf{68.73}  \\  
             \midrule

    \end{tabular}
    }
       
    \label{tab:cross-data}
  
\end{table*}

 

\subsection{Ablation Study}
In this section, we perform ablation studies to demonstrate the effectiveness of each design of the proposed method. 

 \begin{figure}[]
  \centering
  \scalebox{0.9}{ \includegraphics[width=1\linewidth]{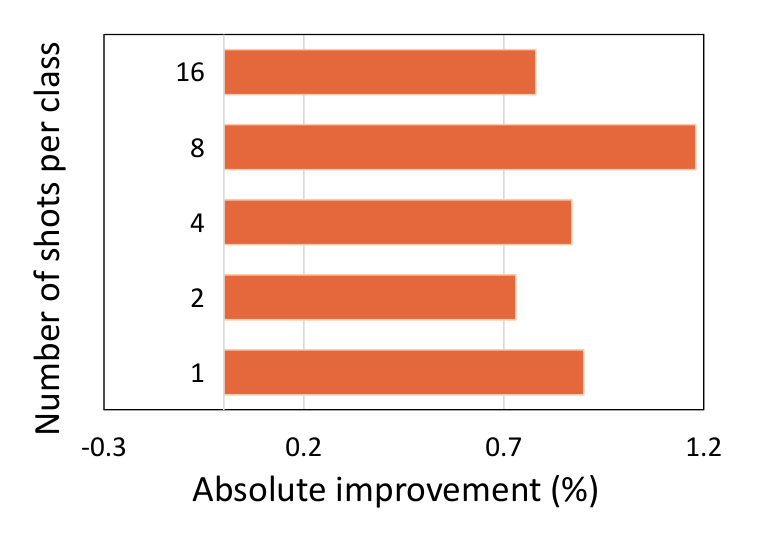}}

   \caption{ The absolute accuracy improvements provided by using \textbf{AVAE} compared to scenarios without \textbf{AVAE}. }

   \label{fig:avae}
\end{figure}

\begin{figure*}[h]
  \centering
 
     \scalebox{0.9}{
        \includegraphics[width=0.99\linewidth]{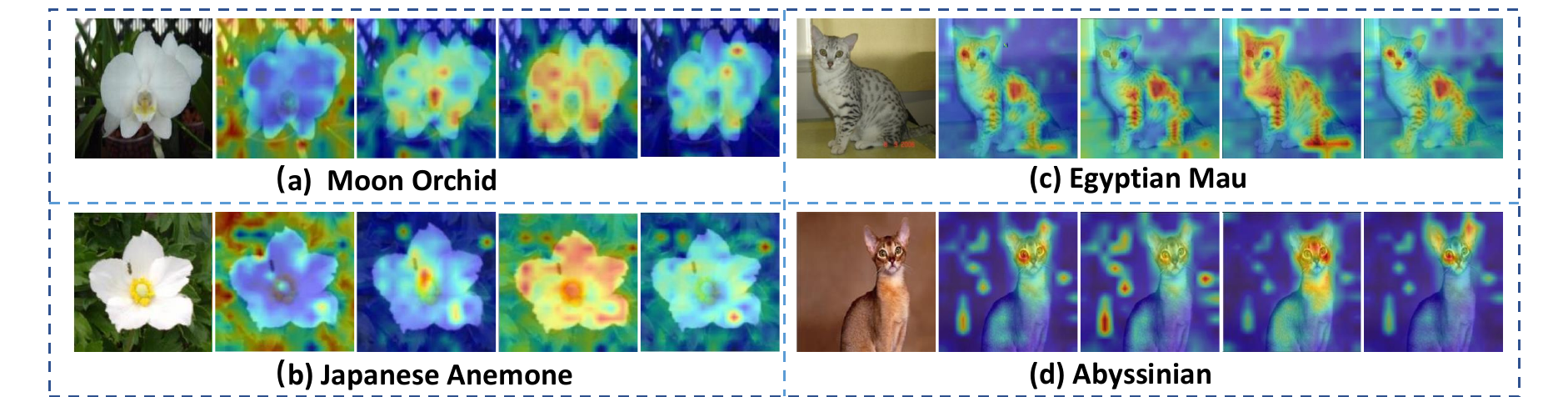}
     
     }

   \caption{The visualization of visual attribute prompts. Guided by textual attribute semantics, visual attribute prompts focus on distinctive visual details, such as the different leaf shapes of the Moon Orchid and Japanese Anemone, the spotted coat of the Egyptian Mau, and the large ears of the Abyssinian.}
 
   \label{fig:visualization}
\end{figure*}

 \begin{figure}[]
  \centering
  \scalebox{0.8}{ \includegraphics[width=1\linewidth]{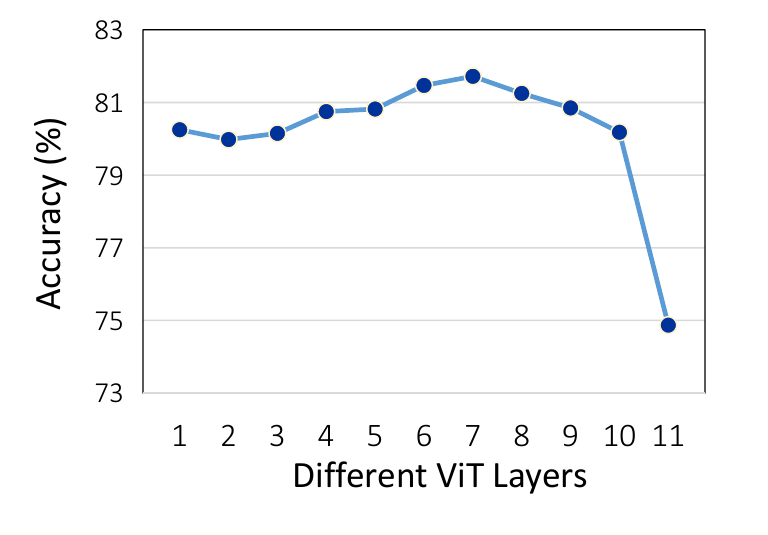}}

   \caption{ The impact of inserting \textbf{AVAE} into different layers of ViT with 1 shot per class. }

   \label{fig:avae2}
\end{figure}

\textbf{Effectiveness of Attribute Prompts.} 
We denote Textual Attribute Prompts as \textbf{TAP} and Visual Attribute Prompts as \textbf{VAP}. We remove TAP and VAP from MAP as our baseline. The results in Table \ref{tab:componets} are analyzed as follows: (1) Compared to the baseline, utilizing TAP powered by the LLM effectively improves the novel accuracy, achieving an accuracy gain of 1.43\%, which demonstrates textual attributes enrich the semantics for novel classes. (2) The incorporation of VAP shows a distinct performance boost on both base (+1.6\%) and novel classes (+2.11\%). This proves that VAP contributes to enhancing fine-grained visual perception ability by capturing visual attributes.

 \begin{figure}[]
  \centering
    
  \scalebox{1.25}{\includegraphics[width=0.8\linewidth]{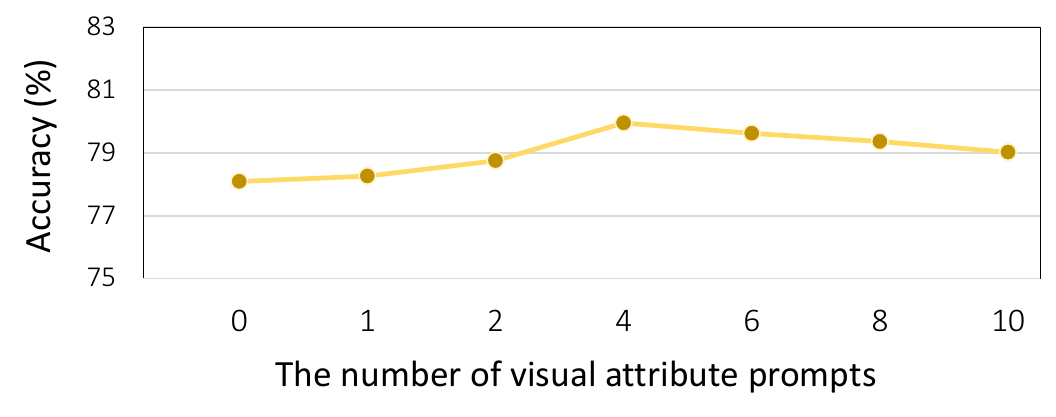}}

   \caption{The impact of the number of visual attribute prompts in the base-to-novel generalization setting.}

   \label{fig:num}
\end{figure} 

\begin{figure}
\centering
\scalebox{1.1}{
\renewcommand{\thefigure}{{\Alph{figure}}}
\includegraphics[width=0.85\linewidth]{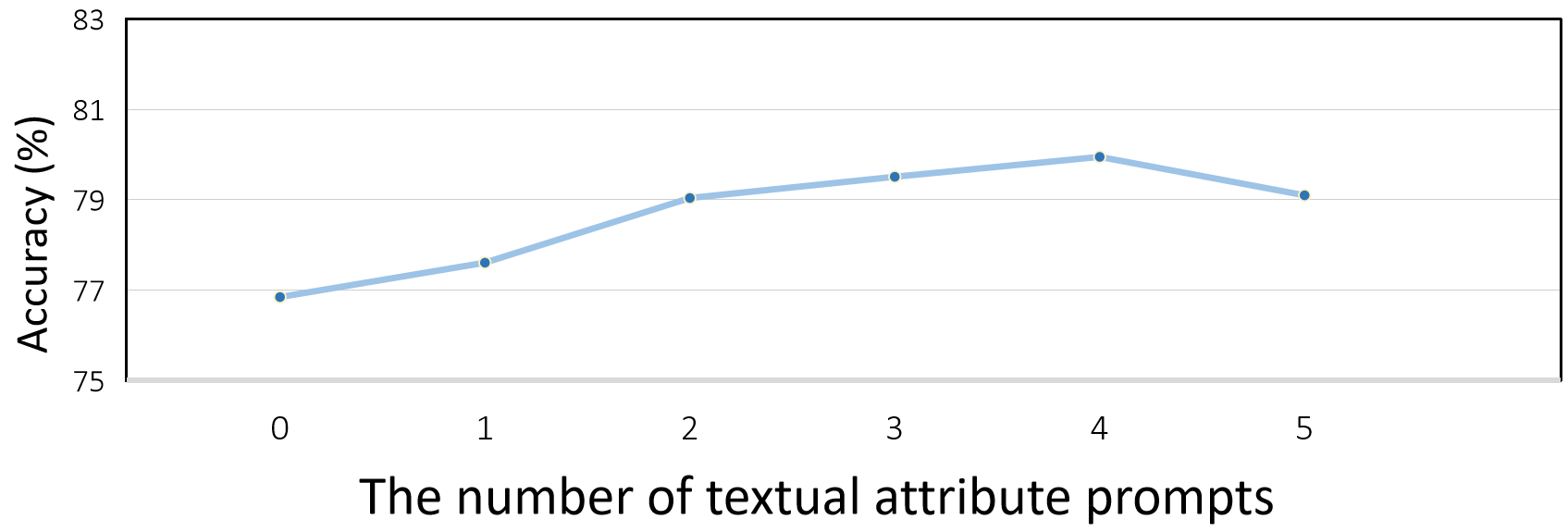}}
\caption{{The impact of the number of textual attribute prompts per class in the base-to-novel generalization setting.}}
\label{fig:t_num}
\end{figure}

\begin{table}[!tb]

    \small \centering
 \setlength{\tabcolsep}{8pt}
  \caption{Ablation results.} 
  
    \scalebox{1.2}[1.2]{
 \begin{tabular}{@{\hspace{10pt}}c@{\hspace{10pt}}c@{\hspace{10pt}}c|c }
        \toprule
            Method & Base & Novel & HM  \\ \hline
         Baseline & 82.20 & 72.22 & 76.41  \\ 
        +TAP(LLM) & 82.06 & 73.65 & 77.36  \\ 
          +TAP+VAP (MAP) & 83.66 & 75.76 & 79.36  \\
   
        \bottomrule
      \end{tabular}

}
       
    \label{tab:componets}
   
\end{table}
\begin{table}[!tb]

    \small \centering
 \setlength{\tabcolsep}{8pt}
  \caption{Complexity results.} 
  
    \scalebox{1.2}[1.2]{
       \begin{tabular}{@{\hspace{10pt}}c|@{\hspace{10pt}}c@{\hspace{10pt}}cc }
        \toprule
             & CoCoOp  & MaPLe  & MAP   \\ \hline
         parameters & 0.04M & 3.56M & 0.74M  \\ 
        GFLOPs & 83.83 & 55.23 & 84.80  \\ 
          test time & 56.70s & 9.58s & 9.79s  \\
   
        \bottomrule
      \end{tabular}
}
       
    \label{tab:flops}
   
\end{table}
\begin{table}[!t]
    \small \centering
 \setlength{\tabcolsep}{8pt}
 \caption{{The impact of using different LLMs.}}
  
    \scalebox{1.2}[1.2]{
 \begin{tabular}{@{\hspace{10pt}}c@{\hspace{10pt}}c@{\hspace{10pt}}c|c }
        \toprule
            Method & Base & Novel & HM  \\ \hline
         Qwen-1.8B-Chat & 97.47 & 73.23 & 83.63  \\ 
        GPT-3.5 & 97.57 & 75.23 & 84.95  \\ 
          Qwen1.5-72B-Chat & 97.77 & 75.30 & 85.08  \\
   
        \bottomrule
      \end{tabular}   
}
    \label{tab:LLMs}
   
\end{table}

\textbf{Effectiveness of Adaptive Visual Attribute Enhancement.} To verify the accuracy improvement when using AVAE, we conduct few-shot image classification experiments on 6 datasets (Flowers102, DTD, UCF101, OxfordPets, Caltech101, Food101). As shown in Figure \ref{fig:avae}, the employment of AVAE brings remarkable performance gains. Furthermore, we investigate the impact of placing AVAE into different ViT layers. As observed from Figure \ref{fig:avae2}, placing AVAE in the middle layers (Layer 6-8) attains superior performance. When applying AVAE in the shallow or deep layers, the performance deteriorates obviously compared to the middle layers. Therefore, the AVAE module should be placed in the middle layers. Initial visual attribute prompts can aggregate visual regional features in shallow layers and continue to capture visual attributes in the remaining layers after enhancement by AVAE.

\textbf{Analysis of Number of Visual Attribute Prompts.} Figure \ref{fig:num} illustrates the averaged harmonic mean accuracy of using varying numbers of visual prompts over 10 datasets in the base-to-novel generalization setting. When the number is as small as 1, the performance gain is quite limited. The accuracy increases with more visual attribute prompts, as more visual attribute characteristics can be captured. However, the accuracy decreases slightly when the number is beyond 4, as an excessive amount of visual attribute prompts may contain redundancy and noises.

{
\textbf{Analysis of Number of Textual Attribute Prompts.} Figure~\ref {fig:t_num} illustrates the averaged harmonic accuracy of using different numbers of textual attribute prompts. According to the experimental results, the introduction of textual attribute prompts indeed improves the performance, demonstrating the effectiveness of textual attribute prompts. The accuracy improves with the incorporation of more textual attribute prompts, as this introduces more descriptive information.
However, when the number of textual attribute prompts exceeds four, the performance decreases. 
This may be attributed to the fact that additional prompts introduce more redundancy. The initial prompts are usually the most relevant and effective, while later ones may include less useful or intuitive descriptions. Increased complexity and less discriminative attributes like size or height can also burden the model, resulting in reduced performance. Overall, the accuracy changes relatively smoothly with different prompt numbers.
}

{
\textbf{Impact of Different LLMs.}
We conduct experiments using other large language models (LLMs), specifically Qwen-1.8B-Chat and Qwen-1.5-72B-Chat~\cite{qwen}, and examine performance variations on the Flowers102 dataset.
The results in Table~\ref{tab:LLMs} show that Qwen-1.5-72B-Chat achieves performance comparable to GPT-3.5. However, when using Qwen-1.8B-Chat, there is a significant performance drop compared to using GPT-3.5 and Qwen-1.5-72B-Chat.
This decline may be attributed to the fact that the outputs from Qwen-1.8B-Chat are sometimes inconsistent, noisy, and occasionally lack meaningful information. These findings suggest that selecting a large language model capable of generating consistent and clear outputs is crucial for maintaining performance.
}

\textbf{Analysis of Complexity.}
We compare different prompting methods about the number of parameters, the GFLOPs, and the test time
 in Table \ref{tab:flops}. MaPLe~\cite{Maple} and MAP enjoy faster inference speeds than CoCoOp~\cite{cocoop}.
Compared with MaPLe, MAP is more parameter-efficient (0.74M vs 3.56M). 
The computation cost (GFLOPs) of MAP is higher, but considering the performance 
improvement, it is acceptable.

\textbf{Visualization of Visual Attribute Prompts.} We visualize visual attribute prompts output by the Vision Transformer in Figure \ref{fig:visualization}. It can be observed that different visual attribute prompts focus on various aspects of the image and highlight distinctive visual details. This visualization demonstrates the capacity of visual attribute prompts to augment the model's fine-grained visual perception ability.


\section{Limitation and Future Work}
We use text attributes directly from GPT without manual filtering. Text attributes may contain noise that may hinder accurate classification, such as attributes with high uncertainty, like colors of toad lilies (white, purple, pink, or yellow). On Flowers102~\cite{flowers}, we manually filter improper attributes, resulting in an improvement of 0.37\% in HM. Filtering improper ones has the potential to improve results. We'll design an automatic filter plan in the future.

\section{Conclusion}
In this paper, we propose a Multi-modal Attribute Prompting method to adapt pre-trained Vision-Language models for downstream few-shot tasks. Our method involves modeling visual attributes to enhance the visual fine-grained perception ability. We establish attribute-level alignment, complementing the global alignment to achieve multi-level robust alignment between images and text categories. Extensive experimental results demonstrate the effectiveness.

\section*{Acknowledgments}
This work was supported by National Defense Basic Scientific Research Program of China (JCKY2020903B002), National Natural Science Foundation of China (62306294), Anhui Provincial Natural Science Foundation (2308085QF222), China Postdoctoral Science Foundation (2023M743385) and Youth Innovation Promotion Association CAS.



 
%

\bibliographystyle{IEEEtran}
\bibliography{reference}

\begin{thebibliography}{10}
\providecommand{\url}[1]{#1}
\csname url@samestyle\endcsname
\providecommand{\newblock}{\relax}
\providecommand{\bibinfo}[2]{#2}
\providecommand{\BIBentrySTDinterwordspacing}{\spaceskip=0pt\relax}
\providecommand{\BIBentryALTinterwordstretchfactor}{4}
\providecommand{\BIBentryALTinterwordspacing}{\spaceskip=\fontdimen2\font plus
\BIBentryALTinterwordstretchfactor\fontdimen3\font minus \fontdimen4\font\relax}
\providecommand{\BIBforeignlanguage}[2]{{%
\expandafter\ifx\csname l@#1\endcsname\relax
\typeout{** WARNING: IEEEtran.bst: No hyphenation pattern has been}%
\typeout{** loaded for the language `#1'. Using the pattern for}%
\typeout{** the default language instead.}%
\else
\language=\csname l@#1\endcsname
\fi
#2}}
\providecommand{\BIBdecl}{\relax}
\BIBdecl

\bibitem{CLIP}
A.~Radford, J.~W. Kim, C.~Hallacy, A.~Ramesh, G.~Goh, S.~Agarwal, G.~Sastry, A.~Askell, P.~Mishkin, J.~Clark \emph{et~al.}, ``Learning transferable visual models from natural language supervision,'' in \emph{International Conference on Machine Learning}.\hskip 1em plus 0.5em minus 0.4em\relax PMLR, 2021, pp. 8748--8763.

\bibitem{ALIGN}
C.~Jia, Y.~Yang, Y.~Xia, Y.-T. Chen, Z.~Parekh, H.~Pham, Q.~Le, Y.-H. Sung, Z.~Li, and T.~Duerig, ``Scaling up visual and vision-language representation learning with noisy text supervision,'' in \emph{International Conference on Machine Learning}.\hskip 1em plus 0.5em minus 0.4em\relax PMLR, 2021, pp. 4904--4916.

\bibitem{mei2022guest}
T.~Mei, J.~J. Corso, G.~Kim, J.~Luo, C.~Shen, and H.~Zhang, ``Guest editorial introduction to the special section on video and language,'' \emph{IEEE Transactions on Circuits and Systems for Video Technology}, vol.~32, no.~1, pp. 1--4, 2022.

\bibitem{zhang2020language}
W.~Zhang, C.~Ma, Q.~Wu, and X.~Yang, ``Language-guided navigation via cross-modal grounding and alternate adversarial learning,'' \emph{IEEE Transactions on Circuits and Systems for Video Technology}, vol.~31, no.~9, pp. 3469--3481, 2020.

\bibitem{wei2024fine}
Z.~Wei, Z.~Zhang, P.~Wu, J.~Wang, P.~Wang, and Y.~Zhang, ``Fine-granularity alignment for text-based person retrieval via semantics-centric visual division,'' \emph{IEEE Transactions on Circuits and Systems for Video Technology}, 2024.

\bibitem{zhu2023esa}
H.~Zhu, C.~Zhang, Y.~Wei, S.~Huang, and Y.~Zhao, ``Esa: External space attention aggregation for image-text retrieval,'' \emph{IEEE Transactions on Circuits and Systems for Video Technology}, 2023.

\bibitem{zhou2024unsupervised}
W.~Zhou and Z.~Zhou, ``Unsupervised domain adaption harnessing vision-language pre-training,'' \emph{IEEE Transactions on Circuits and Systems for Video Technology}, 2024.

\bibitem{lin2024clipose}
X.~Lin, M.~Zhu, R.~Dang, G.~Zhou, S.~Shu, F.~Lin, C.~Liu, and Q.~Chen, ``Clipose: Category-level object pose estimation with pre-trained vision-language knowledge,'' \emph{IEEE Transactions on Circuits and Systems for Video Technology}, 2024.

\bibitem{wang2023tridentcap}
L.~Wang, H.~Qiu, B.~Qiu, F.~Meng, Q.~Wu, and H.~Li, ``Tridentcap: Image-fact-style trident semantic framework for stylized image captioning,'' \emph{IEEE Transactions on Circuits and Systems for Video Technology}, 2023.

\bibitem{vlm-adapt2}
R.~Arandjelovi{\'c}, A.~Andonian, A.~Mensch, O.~J. H{\'e}naff, J.-B. Alayrac, and A.~Zisserman, ``Three ways to improve feature alignment for open vocabulary detection,'' \emph{arXiv preprint arXiv:2303.13518}, 2023.

\bibitem{vlm-adapt3}
P.~Kaul, W.~Xie, and A.~Zisserman, ``Multi-modal classifiers for open-vocabulary object detection,'' in \emph{International Conference on Machine Learning}.\hskip 1em plus 0.5em minus 0.4em\relax PMLR, 2023, pp. 15\,946--15\,969.

\bibitem{3dscene}
S.~Peng, K.~Genova, C.~Jiang, A.~Tagliasacchi, M.~Pollefeys, T.~Funkhouser \emph{et~al.}, ``Openscene: 3d scene understanding with open vocabularies,'' in \emph{Proceedings of the IEEE/CVF Conference on Computer Vision and Pattern Recognition}, 2023, pp. 815--824.

\bibitem{3d-2}
C.~Zhu, W.~Zhang, T.~Wang, X.~Liu, and K.~Chen, ``Object2scene: Putting objects in context for open-vocabulary 3d detection,'' \emph{arXiv preprint arXiv:2309.09456}, 2023.

\bibitem{3d-3}
A.~Takmaz, E.~Fedele, R.~W. Sumner, M.~Pollefeys, F.~Tombari, and F.~Engelmann, ``Openmask3d: Open-vocabulary 3d instance segmentation,'' in \emph{Advances in Neural Information Processing Systems 36: Annual Conference on Neural Information Processing Systems}, 2023.

\bibitem{clip-adapter}
P.~Gao, S.~Geng, R.~Zhang, T.~Ma, R.~Fang, Y.~Zhang, H.~Li, and Y.~Qiao, ``Clip-adapter: Better vision-language models with feature adapters,'' \emph{International Journal of Computer Vision}, vol. 132, no.~2, pp. 581--595, 2024.

\bibitem{CoOp}
K.~Zhou, J.~Yang, C.~C. Loy, and Z.~Liu, ``Learning to prompt for vision-language models,'' \emph{International Journal of Computer Vision}, vol. 130, no.~9, pp. 2337--2348, 2022.

\bibitem{SubPT}
C.~Ma, Y.~Liu, J.~Deng, L.~Xie, W.~Dong, and C.~Xu, ``Understanding and mitigating overfitting in prompt tuning for vision-language models,'' \emph{IEEE Transactions on Circuits and Systems for Video Technology}, 2023.

\bibitem{cocoop}
K.~Zhou, J.~Yang, C.~C. Loy, and Z.~Liu, ``Conditional prompt learning for vision-language models,'' in \emph{Proceedings of the IEEE/CVF Conference on Computer Vision and Pattern Recognition}, 2022, pp. 16\,816--16\,825.

\bibitem{dapt}
E.~Cho, J.~Kim, and H.~J. Kim, ``Distribution-aware prompt tuning for vision-language models,'' in \emph{Proceedings of the IEEE/CVF International Conference on Computer Vision}, 2023, pp. 22\,004--22\,013.

\bibitem{PromptSRC}
M.~U. Khattak, S.~T. Wasim, M.~Naseer, S.~Khan, M.-H. Yang, and F.~S. Khan, ``Self-regulating prompts: Foundational model adaptation without forgetting,'' in \emph{Proceedings of the IEEE/CVF International Conference on Computer Vision}, 2023, pp. 15\,190--15\,200.

\bibitem{ProDa}
Y.~Lu, J.~Liu, Y.~Zhang, Y.~Liu, and X.~Tian, ``Prompt distribution learning,'' in \emph{Proceedings of the IEEE/CVF Conference on Computer Vision and Pattern Recognition}, 2022, pp. 5206--5215.

\bibitem{rpo}
D.~Lee, S.~Song, J.~Suh, J.~Choi, S.~Lee, and H.~J. Kim, ``Read-only prompt optimization for vision-language few-shot learning,'' in \emph{Proceedings of the IEEE/CVF International Conference on Computer Vision}, 2023, pp. 1401--1411.

\bibitem{Maple}
M.~U. Khattak, H.~Rasheed, M.~Maaz, S.~Khan, and F.~S. Khan, ``Maple: Multi-modal prompt learning,'' in \emph{Proceedings of the IEEE/CVF Conference on Computer Vision and Pattern Recognition}, 2023, pp. 19\,113--19\,122.

\bibitem{Multi-des}
Z.~Feng, A.~Bair, and J.~Z. Kolter, ``Leveraging multiple descriptive features for robust few-shot image learning,'' \emph{arXiv preprint arXiv:2307.04317}, 2023.

\bibitem{classification_via_des}
S.~Menon and C.~Vondrick, ``Visual classification via description from large language models,'' in \emph{International Conference on Learning Representations,}, 2023.

\bibitem{gpt_enhance_clip}
M.~Maniparambil, C.~Vorster, D.~Molloy, N.~Murphy, K.~McGuinness, and N.~E. O'Connor, ``Enhancing clip with gpt-4: Harnessing visual descriptions as prompts,'' in \emph{Proceedings of the IEEE/CVF International Conference on Computer Vision}, 2023, pp. 262--271.

\bibitem{gpt3}
T.~Brown, B.~Mann, N.~Ryder, M.~Subbiah, J.~D. Kaplan, P.~Dhariwal, A.~Neelakantan, P.~Shyam, G.~Sastry, A.~Askell \emph{et~al.}, ``Language models are few-shot learners,'' \emph{Advances in Neural Information Processing Systems}, vol.~33, pp. 1877--1901, 2020.

\bibitem{gpt4}
R.~OpenAI, ``Gpt-4 technical report. arxiv 2303.08774,'' \emph{View in Article}, 2023.

\bibitem{llmsurvey}
W.~X. Zhao, K.~Zhou, J.~Li, T.~Tang, X.~Wang, Y.~Hou, Y.~Min, B.~Zhang, J.~Zhang, Z.~Dong \emph{et~al.}, ``A survey of large language models,'' \emph{arXiv preprint arXiv:2303.18223}, 2023.

\bibitem{OT}
C.~Villani, \emph{Optimal transport: old and new}.\hskip 1em plus 0.5em minus 0.4em\relax Springer, 2009, vol. 338.

\bibitem{sinkhorn}
M.~Cuturi, ``Sinkhorn distances: Lightspeed computation of optimal transport,'' \emph{Advances in neural information processing systems}, vol.~26, 2013.

\bibitem{yu2019multimodal}
J.~Yu, J.~Li, Z.~Yu, and Q.~Huang, ``Multimodal transformer with multi-view visual representation for image captioning,'' \emph{IEEE transactions on circuits and systems for video technology}, vol.~30, no.~12, pp. 4467--4480, 2019.

\bibitem{yang2020grounding}
Z.~Yang, T.~Kumar, T.~Chen, J.~Su, and J.~Luo, ``Grounding-tracking-integration,'' \emph{IEEE Transactions on Circuits and Systems for Video Technology}, vol.~31, no.~9, pp. 3433--3443, 2020.

\bibitem{flava}
A.~Singh, R.~Hu, V.~Goswami, G.~Couairon, W.~Galuba, M.~Rohrbach, and D.~Kiela, ``Flava: A foundational language and vision alignment model,'' in \emph{Proceedings of the IEEE/CVF Conference on Computer Vision and Pattern Recognition}, 2022, pp. 15\,638--15\,650.

\bibitem{lit}
X.~Zhai, X.~Wang, B.~Mustafa, A.~Steiner, D.~Keysers, A.~Kolesnikov, and L.~Beyer, ``Lit: Zero-shot transfer with locked-image text tuning,'' in \emph{Proceedings of the IEEE/CVF Conference on Computer Vision and Pattern Recognition}, 2022, pp. 18\,123--18\,133.

\bibitem{Florence}
L.~Yuan, D.~Chen, Y.-L. Chen, N.~Codella, X.~Dai, J.~Gao, H.~Hu, X.~Huang, B.~Li, C.~Li \emph{et~al.}, ``Florence: A new foundation model for computer vision,'' \emph{arXiv preprint arXiv:2111.11432}, 2021.

\bibitem{jiang2020multi-metric-learning}
W.~Jiang, K.~Huang, J.~Geng, and X.~Deng, ``Multi-scale metric learning for few-shot learning,'' \emph{IEEE Transactions on Circuits and Systems for Video Technology}, vol.~31, no.~3, pp. 1091--1102, 2020.

\bibitem{cheng2021meta-learning}
M.~Cheng, H.~Wang, and Y.~Long, ``Meta-learning-based incremental few-shot object detection,'' \emph{IEEE Transactions on Circuits and Systems for Video Technology}, vol.~32, no.~4, pp. 2158--2169, 2021.

\bibitem{wang2023few}
X.~Wang, X.~Wang, B.~Jiang, and B.~Luo, ``Few-shot learning meets transformer: Unified query-support transformers for few-shot classification,'' \emph{IEEE Transactions on Circuits and Systems for Video Technology}, 2023.

\bibitem{xu2022gct_few}
R.~Xu, L.~Xing, S.~Shao, L.~Zhao, B.~Liu, W.~Liu, and Y.~Zhou, ``Gct: Graph co-training for semi-supervised few-shot learning,'' \emph{IEEE Transactions on Circuits and Systems for Video Technology}, vol.~32, no.~12, pp. 8674--8687, 2022.

\bibitem{zhang2022mfnet_few}
M.~Zhang, M.~Shi, and L.~Li, ``Mfnet: Multiclass few-shot segmentation network with pixel-wise metric learning,'' \emph{IEEE Transactions on Circuits and Systems for Video Technology}, vol.~32, no.~12, pp. 8586--8598, 2022.

\bibitem{zhang2019few}
C.~Zhang, C.~Li, and J.~Cheng, ``Few-shot visual classification using image pairs with binary transformation,'' \emph{IEEE Transactions on Circuits and Systems for Video Technology}, vol.~30, no.~9, pp. 2867--2871, 2019.

\bibitem{dang2023counterfactual_few}
Z.~Dang, M.~Luo, C.~Jia, C.~Yan, X.~Chang, and Q.~Zheng, ``Counterfactual generation framework for few-shot learning,'' \emph{IEEE Transactions on Circuits and Systems for Video Technology}, 2023.

\bibitem{nlp_p3}
Z.~Jiang, F.~F. Xu, J.~Araki, and G.~Neubig, ``How can we know what language models know?'' \emph{Transactions of the Association for Computational Linguistics}, vol.~8, pp. 423--438, 2020.

\bibitem{nlp_p4_tune}
X.~L. Li and P.~Liang, ``Prefix-tuning: Optimizing continuous prompts for generation,'' in \emph{Proceedings of the 59th Annual Meeting of the Association for Computational Linguistics and the 11th International Joint Conference on Natural Language Processing}, 2021, pp. 4582--4597.

\bibitem{nlp_p5_tune}
B.~Lester, R.~Al{-}Rfou, and N.~Constant, ``The power of scale for parameter-efficient prompt tuning,'' in \emph{Proceedings of the 2021 Conference on Empirical Methods in Natural Language Processing,}, pp. 3045--3059.

\bibitem{nlp_p6_ppt}
Y.~Gu, X.~Han, Z.~Liu, and M.~Huang, ``{PPT:} pre-trained prompt tuning for few-shot learning,'' in \emph{Proceedings of the 60th Annual Meeting of the Association for Computational Linguistics}, 2022, pp. 8410--8423.

\bibitem{nlp_p7gpt}
X.~Liu, Y.~Zheng, Z.~Du, M.~Ding, Y.~Qian, Z.~Yang, and J.~Tang, ``Gpt understands, too,'' \emph{AI Open}, 2023.

\bibitem{vpt}
M.~Jia, L.~Tang, B.-C. Chen, C.~Cardie, S.~Belongie, B.~Hariharan, and S.-N. Lim, ``Visual prompt tuning,'' in \emph{European Conference on Computer Vision}.\hskip 1em plus 0.5em minus 0.4em\relax Springer, 2022, pp. 709--727.

\bibitem{vit}
A.~Dosovitskiy, L.~Beyer, A.~Kolesnikov, D.~Weissenborn, X.~Zhai, T.~Unterthiner, M.~Dehghani, M.~Minderer, G.~Heigold, S.~Gelly, J.~Uszkoreit, and N.~Houlsby, ``An image is worth 16x16 words: Transformers for image recognition at scale,'' in \emph{International Conference on Learning Representations, {ICLR} 2021}.

\bibitem{Learning_visual_attributes_7}
V.~Ferrari and A.~Zisserman, ``Learning visual attributes,'' \emph{Advances in neural information processing systems}, vol.~20, 2007.

\bibitem{visual_attribute_search_17}
N.~Kumar, A.~Berg, P.~N. Belhumeur, and S.~Nayar, ``Describable visual attributes for face verification and image search,'' \emph{IEEE Transactions on Pattern Analysis and Machine Intelligence}, vol.~33, no.~10, pp. 1962--1977, 2011.

\bibitem{visual_attribute_regconition_37}
S.~Wang, Z.~Wang, H.~Li, J.~Chang, W.~Ouyang, and Q.~Tian, ``Accurate fine-grained object recognition with structure-driven relation graph networks,'' \emph{International Journal of Computer Vision}, vol. 132, no.~1, pp. 137--160, 2024.

\bibitem{visual_attribute_scene_25}
G.~Patterson, C.~Xu, H.~Su, and J.~Hays, ``The sun attribute database: Beyond categories for deeper scene understanding,'' \emph{International Journal of Computer Vision}, vol. 108, pp. 59--81, 2014.

\bibitem{visual_attribute_10}
J.~Huang, R.~S. Feris, Q.~Chen, and S.~Yan, ``Cross-domain image retrieval with a dual attribute-aware ranking network,'' in \emph{Proceedings of the IEEE international conference on computer vision}, 2015, pp. 1062--1070.

\bibitem{visual_attribute_49}
H.~Zhang, X.~Cao, and R.~Wang, ``Audio visual attribute discovery for fine-grained object recognition,'' in \emph{Proceedings of the AAAI Conference on Artificial Intelligence}, vol.~32, no.~1, 2018.

\bibitem{visual_attribute_43}
X.-S. Wei, Y.~Shen, X.~Sun, H.-J. Ye, and J.~Yang, ``Learning attribute-aware hash codes for large-scale fine-grained image retrieval,'' \emph{Advances in Neural Information Processing Systems}, vol.~34, pp. 5720--5730, 2021.

\bibitem{VAPNet}
S.~Wang, J.~Chang, H.~Li, Z.~Wang, W.~Ouyang, and Q.~Tian, ``Learning to parameterize visual attributes for open-set fine-grained retrieval,'' \emph{Advances in Neural Information Processing Systems}, vol.~36, 2024.

\bibitem{contra}
P.~Khosla, P.~Teterwak, C.~Wang, A.~Sarna, Y.~Tian, P.~Isola, A.~Maschinot, C.~Liu, and D.~Krishnan, ``Supervised contrastive learning,'' \emph{Advances in Neural Information Processing Systems}, vol.~33, pp. 18\,661--18\,673, 2020.

\bibitem{2018bert}
J.~Devlin, M.-W. Chang, K.~Lee, and K.~Toutanova, ``Bert: Pre-training of deep bidirectional transformers for language understanding,'' \emph{arXiv preprint arXiv:1810.04805}, 2018.

\bibitem{ViT_see_like}
M.~Raghu, T.~Unterthiner, S.~Kornblith, C.~Zhang, and A.~Dosovitskiy, ``Do vision transformers see like convolutional neural networks?'' \emph{Advances in neural information processing systems}, vol.~34, pp. 12\,116--12\,128, 2021.

\bibitem{fromcliptodino}
D.~Jiang, Y.~Liu, S.~Liu, X.~Zhang, J.~Li, H.~Xiong, and Q.~Tian, ``From clip to dino: Visual encoders shout in multi-modal large language models,'' 2023.

\bibitem{HM}
Y.~Xian, B.~Schiele, and Z.~Akata, ``Zero-shot learning-the good, the bad and the ugly,'' in \emph{Proceedings of the IEEE Conference on Computer Vision and Pattern Recognition}, 2017, pp. 4582--4591.

\bibitem{food101}
L.~Bossard, M.~Guillaumin, and L.~Van~Gool, ``Food-101--mining discriminative components with random forests,'' in \emph{European Conference on Computer Vision}.\hskip 1em plus 0.5em minus 0.4em\relax Springer, 2014, pp. 446--461.

\bibitem{DTD}
M.~Cimpoi, S.~Maji, I.~Kokkinos, S.~Mohamed, and A.~Vedaldi, ``Describing textures in the wild,'' in \emph{Proceedings of the IEEE Conference on Computer Vision and Pattern Recognition}, 2014, pp. 3606--3613.

\bibitem{imagenet}
J.~Deng, W.~Dong, R.~Socher, L.-J. Li, K.~Li, and L.~Fei-Fei, ``Imagenet: A large-scale hierarchical image database,'' in \emph{Proceedings of the IEEE Conference on Computer Vision and Pattern Recognition}.\hskip 1em plus 0.5em minus 0.4em\relax Ieee, 2009, pp. 248--255.

\bibitem{cal}
L.~Fei-Fei, R.~Fergus, and P.~Perona, ``Learning generative visual models from few training examples: An incremental bayesian approach tested on 101 object categories,'' in \emph{Proceedings of the IEEE Conference on Computer Vision and Pattern Recognition}.\hskip 1em plus 0.5em minus 0.4em\relax IEEE, 2004, pp. 178--178.

\bibitem{eurosat}
P.~Helber, B.~Bischke, A.~Dengel, and D.~Borth, ``Eurosat: A novel dataset and deep learning benchmark for land use and land cover classification,'' \emph{IEEE Journal of Selected Topics in Applied Earth Observations and Remote Sensing}, vol.~12, no.~7, pp. 2217--2226, 2019.

\bibitem{cars}
J.~Krause, M.~Stark, J.~Deng, and L.~Fei-Fei, ``3d object representations for fine-grained categorization,'' in \emph{Proceedings of the IEEE International Conference on Computer Vision Workshops}, 2013, pp. 554--561.

\bibitem{aircraft}
S.~Maji, E.~Rahtu, J.~Kannala, M.~Blaschko, and A.~Vedaldi, ``Fine-grained visual classification of aircraft,'' \emph{arXiv preprint arXiv:1306.5151}, 2013.

\bibitem{flowers}
M.-E. Nilsback and A.~Zisserman, ``Automated flower classification over a large number of classes,'' in \emph{Indian Conference on Computer Vision, Graphics \& Image processing}.\hskip 1em plus 0.5em minus 0.4em\relax IEEE, 2008, pp. 722--729.

\bibitem{pets}
O.~M. Parkhi, A.~Vedaldi, A.~Zisserman, and C.~Jawahar, ``Cats and dogs,'' in \emph{Proceedings of the IEEE Conference on Computer Vision and Pattern Recognition}.\hskip 1em plus 0.5em minus 0.4em\relax IEEE, 2012, pp. 3498--3505.

\bibitem{sun}
J.~Xiao, J.~Hays, K.~A. Ehinger, A.~Oliva, and A.~Torralba, ``Sun database: Large-scale scene recognition from abbey to zoo,'' in \emph{Proceedings of the IEEE Conference on Computer Vision and Pattern Recognition}.\hskip 1em plus 0.5em minus 0.4em\relax IEEE, 2010, pp. 3485--3492.

\bibitem{imagenet-r}
D.~Hendrycks, S.~Basart, N.~Mu, S.~Kadavath, F.~Wang, E.~Dorundo, R.~Desai, T.~Zhu, S.~Parajuli, M.~Guo \emph{et~al.}, ``The many faces of robustness: A critical analysis of out-of-distribution generalization,'' in \emph{Proceedings of the IEEE/CVF International Conference on Computer Vision}, 2021, pp. 8340--8349.

\bibitem{imagenet-a}
D.~Hendrycks, K.~Zhao, S.~Basart, J.~Steinhardt, and D.~Song, ``Natural adversarial examples,'' in \emph{Proceedings of the IEEE/CVF Conference on Computer Vision and Pattern Recognition}, 2021, pp. 15\,262--15\,271.

\bibitem{imagenetV2}
B.~Recht, R.~Roelofs, L.~Schmidt, and V.~Shankar, ``Do imagenet classifiers generalize to imagenet?'' in \emph{International Conference on Machine Learning}.\hskip 1em plus 0.5em minus 0.4em\relax PMLR, 2019, pp. 5389--5400.

\bibitem{imagenet-s}
H.~Wang, S.~Ge, Z.~Lipton, and E.~P. Xing, ``Learning robust global representations by penalizing local predictive power,'' \emph{Advances in Neural Information Processing Systems}, vol.~32, 2019.

\bibitem{Tip-adapter}
R.~Zhang, W.~Zhang, R.~Fang, P.~Gao, K.~Li, J.~Dai, Y.~Qiao, and H.~Li, ``Tip-adapter: Training-free adaption of clip for few-shot classification,'' in \emph{European conference on computer vision}.\hskip 1em plus 0.5em minus 0.4em\relax Springer, 2022, pp. 493--510.

\bibitem{qwen}
J.~B. et~al., ``Qwen technical report,'' \emph{arXiv preprint arXiv:2309.16609}, 2023.

\end{thebibliography}


 

\begin{IEEEbiography}[{\includegraphics[width=1in,height=1.25in,clip,keepaspectratio]{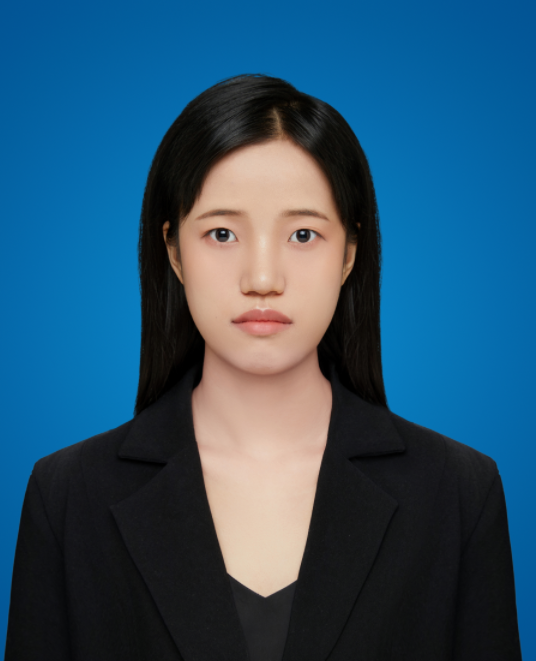}}]{Xin Liu}
received a bachelor's degree in Information Security from the University of Science and Technology of China in 2022. She is now pursuing a master degree in Control Science and Engineering at University of Science and Technology of China. Her research interests include computer vision and deep learning, especially few-shot learning and multi-modal learning.
\end{IEEEbiography}

\begin{IEEEbiography}[{\includegraphics[width=1in,height=1.25in,clip,keepaspectratio]{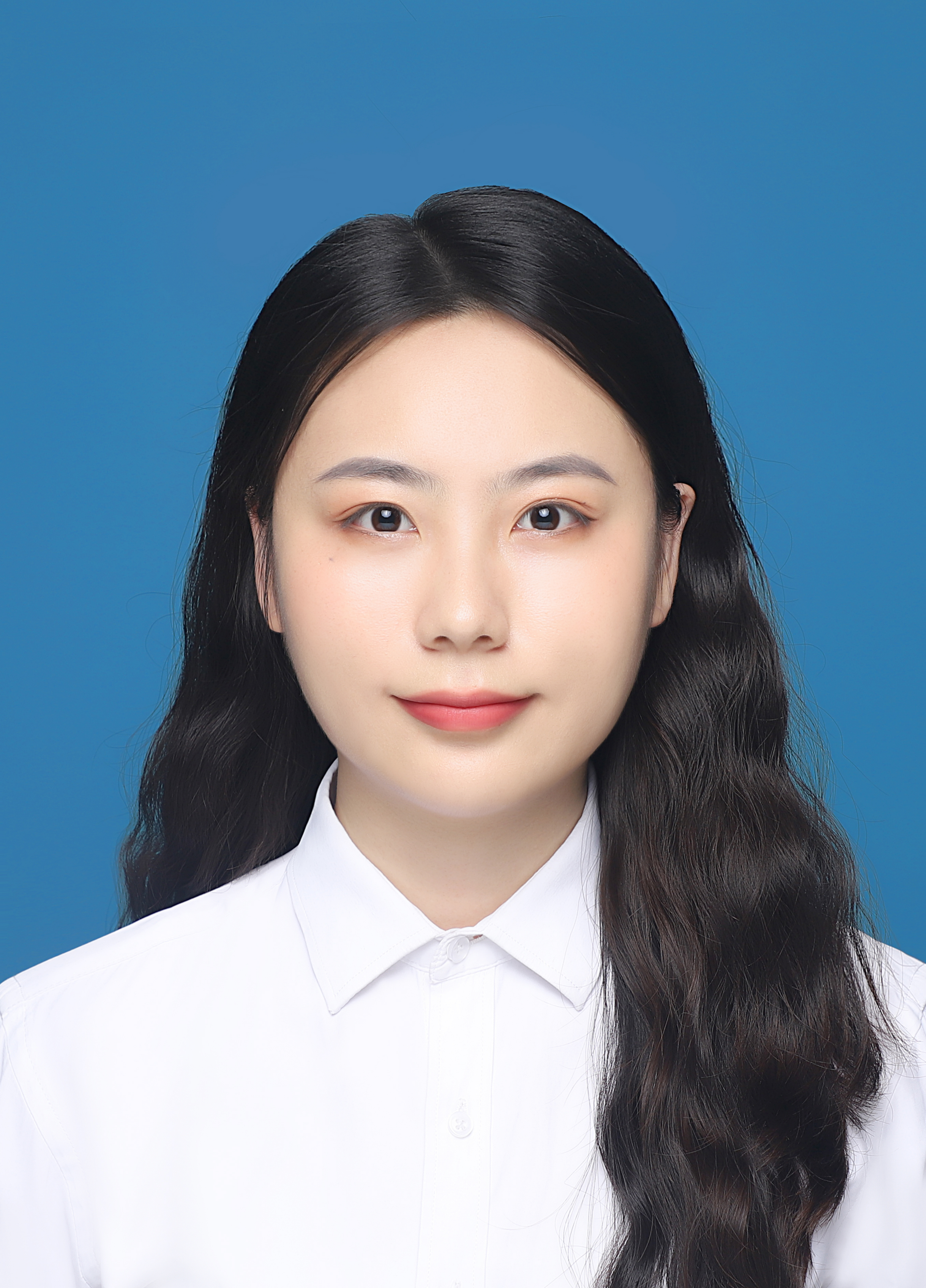}}]{Jiamin Wu}
received the bachelor’s degree in the School of Electronic Engineering, Xidian University, Xian, Shaanxi, China. She is studying for her doctorate in the Department of Automation, University of Science and Technology of China, Hefei, Anhui, China. Her research interests include pattern recognition, computer vision and deep learning. She is currently focusing on zero-shot and few-shot learning.

\end{IEEEbiography}

\begin{IEEEbiography}[{\includegraphics[width=1in,height=1.25in,clip,keepaspectratio]{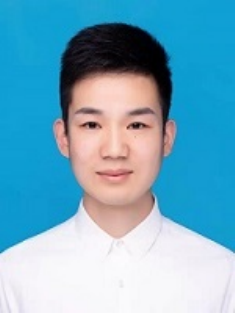}}]{Wenfei Yang}
received the bachelor's degree in
Electronic Engineering and Information Science in
2017, and the Ph.D. degree in pattern recognition
and intelligent systems from the department of Automation,
University of Science and Technology of China, Hefei, China, in 2022.
Currently, he is a post-doctor in Control Science and Engineering, University of Science and Technology of China.
His current research interests include computer vision
and machine learning, especially action detection and object detection.\\
\end{IEEEbiography}

\begin{IEEEbiography}[{\includegraphics[width=1in,height=1.25in,clip,keepaspectratio]{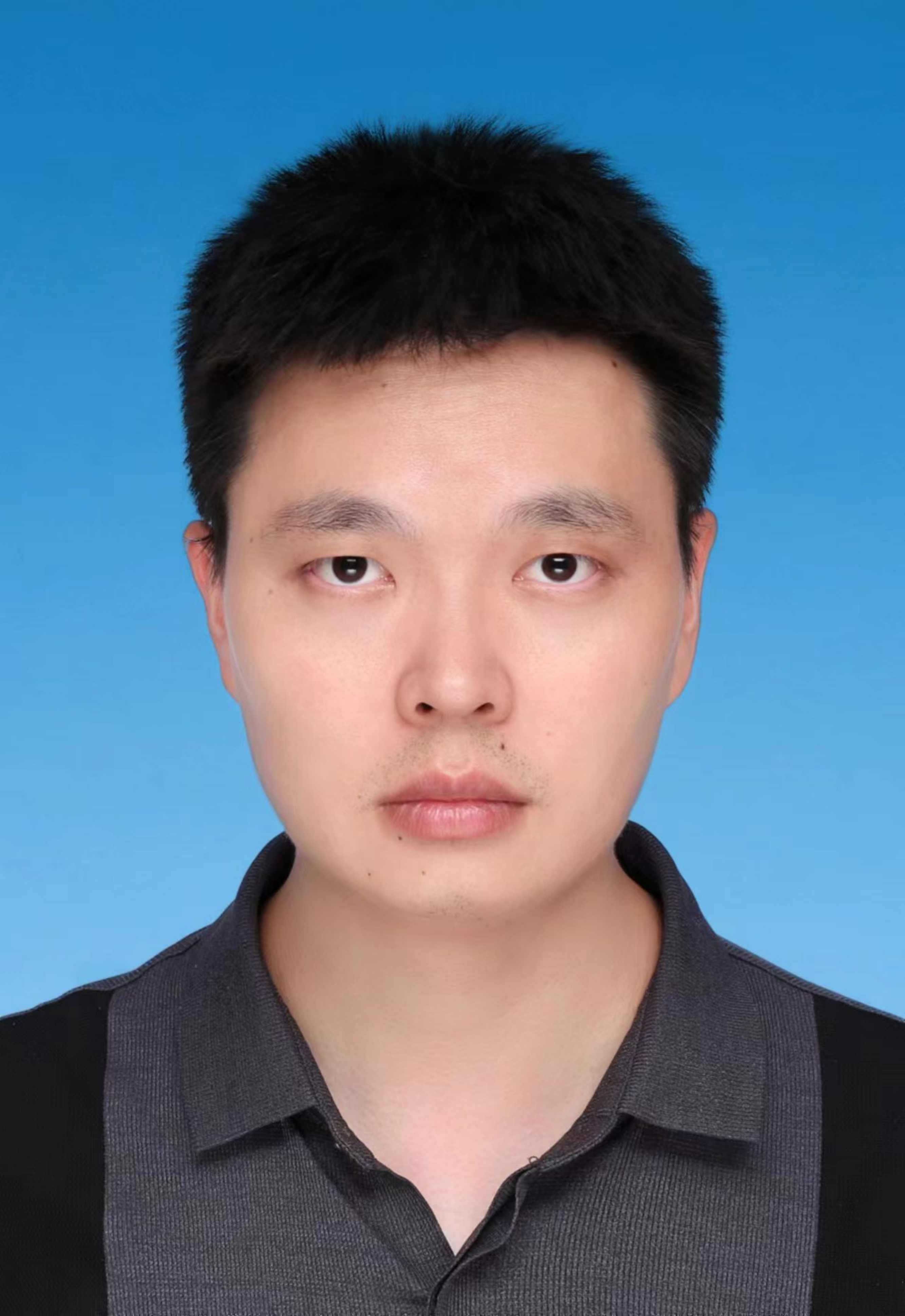}}]{Xu Zhou}
received the PhD degree in computer science and technology from Huazhong University of Science and Technology in 2016. His research interests span the areas of large language model, NLP system design and reinforcement learning.

\end{IEEEbiography}
\begin{IEEEbiography}[{\includegraphics[width=1in,height=1.25in,clip,keepaspectratio]{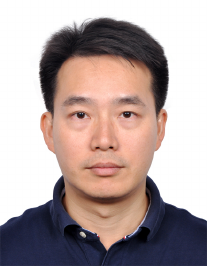}}]{Tianzhu Zhang}
received the bachelor’s degree in communications and information technology from Beijing Institute of Technology, Beijing, China, in 2006, and the Ph.D. degree in pattern recognition and intelligent systems from the Institute of Automation, Chinese Academy of Sciences, Beijing, China, in 2011. Currently, he is a Professor at the Department of Automation, University of Science and Technology of China, Hefei, Anhui, China. His current research interests include computer vision and multimedia, especially action recognition, object classification, object tracking, and social event analysis.

\end{IEEEbiography}

\vfill

\end{document}